\documentclass{article} 
\usepackage{iclr2027_conference,times}
\usepackage[T1]{fontenc}

\usepackage{amsmath,amsfonts,bm}

\def\eqref#1{equation~\ref{#1}}

\def\1{\bm{1}}

\DeclareMathAlphabet{\mathsfit}{\encodingdefault}{\sfdefault}{m}{sl}
\SetMathAlphabet{\mathsfit}{bold}{\encodingdefault}{\sfdefault}{bx}{n}

\usepackage{hyperref}
\usepackage{url}
\usepackage{graphicx}
\usepackage{booktabs}
\usepackage{amssymb}
\usepackage{xcolor}
\usepackage{colortbl}
\definecolor{BestGreen}{HTML}{E8F3E8}
\usepackage{longtable}
\usepackage{wrapfig}
\usepackage{listings}
\usepackage{tabularx}
\usepackage{tcolorbox}
\tcbuselibrary{listings,breakable,skins}
\makeatletter
\newenvironment{papertable}{\par\medskip\noindent\begin{minipage}{\linewidth}\def\@captype{table}\centering}{\end{minipage}\par\medskip}
\makeatother
\newcolumntype{Y}{>{\centering\arraybackslash}X}
\newcolumntype{V}{>{\hsize=.9\hsize\linewidth=\hsize\centering\arraybackslash}X}
\newcolumntype{F}{>{\hsize=1.1\hsize\linewidth=\hsize\centering\arraybackslash}X}

\newlength{\MECoMeanBox}
\newlength{\MECoSDBlock}
\newcommand{\MECoStat}[4]{\begingroup\settowidth{\MECoMeanBox}{$\mathbf{#1}$}\settowidth{\MECoSDBlock}{$#2$}\mbox{\makebox[\MECoMeanBox][r]{$#3$}\,\if\relax\detokenize{#4}\relax\phantom{$\pm$}\makebox[\MECoSDBlock]{}\else$\pm$\makebox[\MECoSDBlock][r]{$#4$}\fi}\endgroup}
\newcommand{\MECoNum}[2]{\begingroup\settowidth{\MECoMeanBox}{$\mathbf{#1}$}\makebox[\MECoMeanBox][r]{$#2$}\endgroup}

\newcommand{\MECoStack}[2]{\if\relax\detokenize{#2}\relax$#1$\else\shortstack[r]{$#1$\\$\pm #2$}\fi}
\newcommand{\MECoDepthStat}[3]{\begingroup\settowidth{\MECoMeanBox}{$\pm\mathbf{#1}$}\if\relax\detokenize{#3}\relax\makebox[\MECoMeanBox][r]{$#2$}\else\shortstack[r]{\makebox[\MECoMeanBox][r]{$#2$}\\\makebox[\MECoMeanBox][r]{$\pm #3$}}\fi\endgroup}

\definecolor{PromptGreen}{HTML}{307357}
\definecolor{PromptBackground}{HTML}{F0F8F4}
\definecolor{CodeBackground}{HTML}{F5F7FA}
\definecolor{CodeBorder}{HTML}{8FAAC9}
\definecolor{CodeKeyword}{HTML}{174D7A}
\definecolor{CodeComment}{HTML}{53758B}
\definecolor{CodeString}{HTML}{346B65}

\newtcblisting{promptbox}[1]{
 enhanced,
 breakable=false,
 listing only,
 colback=PromptBackground,
 colframe=PromptGreen,
 boxrule=0.45pt,
 arc=2pt,
 left=8pt,right=8pt,top=10pt,bottom=7pt,
 before skip=12pt,after skip=10pt,
 title={#1},
 fonttitle=\bfseries\small,
 coltitle=white,
 attach boxed title to top left={xshift=7pt,yshift=-2mm},
 boxed title style={colback=PromptGreen,colframe=PromptGreen,boxrule=0pt,arc=1pt,left=5pt,right=5pt,top=3pt,bottom=3pt},
 listing options={
  language={},
  basicstyle=\ttfamily\fontsize{7.5}{10}\selectfont,
  columns=fullflexible,keepspaces=true,
  breaklines=true,breakatwhitespace=false,
  breakautoindent=false,breakindent=0pt,
  showstringspaces=false,numbers=none,
  aboveskip=0pt,belowskip=0pt,
  keywordstyle=\ttfamily,commentstyle=\ttfamily,stringstyle=\ttfamily
 }
}

\lstdefinestyle{PaperCode}{
 language=Python,
 basicstyle=\ttfamily\fontsize{7}{8.2}\selectfont,
 keywordstyle=\color{CodeKeyword}\bfseries,
 commentstyle=\color{CodeComment},
 stringstyle=\color{CodeString},
 backgroundcolor=\color{CodeBackground},
 rulecolor=\color{CodeBorder},
 frame=single,framerule=0.35pt,framesep=6pt,
 numbers=left,stepnumber=1,firstnumber=1,
 numberstyle=\ttfamily\fontsize{6}{7}\selectfont\color{CodeComment},
 numbersep=8pt,
 xleftmargin=22pt,xrightmargin=7pt,
 framexleftmargin=18pt,framexrightmargin=0pt,
 columns=fullflexible,keepspaces=true,showstringspaces=false,
 breaklines=true,breakatwhitespace=false,breakindent=14pt,
 tabsize=4,
 aboveskip=10pt,belowskip=12pt
}

\title{Multi-Task Evolution for Zero-Shot \\ Cross-Problem Generalization using LLMs}

\author{
Zhuoliang Xie$^{1,2}$,
Changliang Zhou$^{1,2}$,
Genghui Li$^{3}$,
Zhenkun Wang$^{1,2}$\thanks{Corresponding author}\\
$^1$School of Automation and Intelligent Manufacturing, \\
\ \ Southern University of Science and Technology, Shenzhen, China \\
$^2$Guangdong Provincial Key Laboratory of Fully Actuated System Control Theory and Technology,\\
\ \ Southern University of Science and Technology, Shenzhen, China \\
$^3$ College of Computer Science and Software Engineering, Shenzhen University, Shenzhen, China\\
  \texttt{\{xiezl2025, zhoucl2022\}@mail.sustech.edu.cn}  \\
  \texttt{ligh@szu.edu.cn}, 
\texttt{wangzk3@sustech.edu.cn} \\
}

\newcommand{\method}{MECo}
\newcounter{mugealgorithm}

\iclrfinalcopy 
\begin{document}
\maketitle

\begin{abstract}
Designing effective heuristics for diverse combinatorial optimization problems requires substantial expertise and repeated search. Large language models (LLMs) automate heuristic generation and refinement, but heuristic search typically depends on evaluation feedback from the problem being optimized. Generalizing to new problem definitions using only source-task feedback therefore remains a central challenge. We introduce MECo, an LLM-driven multi-task evolutionary framework for zero-shot cross-problem generalization. MECo maintains task-conditioned heuristic populations and uses a transfer gap based on cross-task population performance to guide their interactions. These interactions enable the transfer and recombination of heuristics. A complementary selection criterion then constructs a compact heuristic set by rewarding each member's additional coverage of source combinations. The selected set is applied to target problems without further search or adaptation. Experiments on 32 problem variants across vehicle routing (VRP) and flexible job-shop scheduling (FJSP) show that MECo achieves the lowest mean costs compared with eight automated heuristic design (AHD) baselines under the same budgets. On out-of-domain problems, it outperforms the strongest baseline in each family. Moreover, integrating the framework of MECo with different AHD methods improves their ID and OOD performance in both families, supporting its effectiveness across different methods.
\end{abstract}

\section{Introduction}
\label{sec:introduction}

\begin{wrapfigure}[19]{r}{0.6\textwidth}
 \centering
 \includegraphics[width=\linewidth]{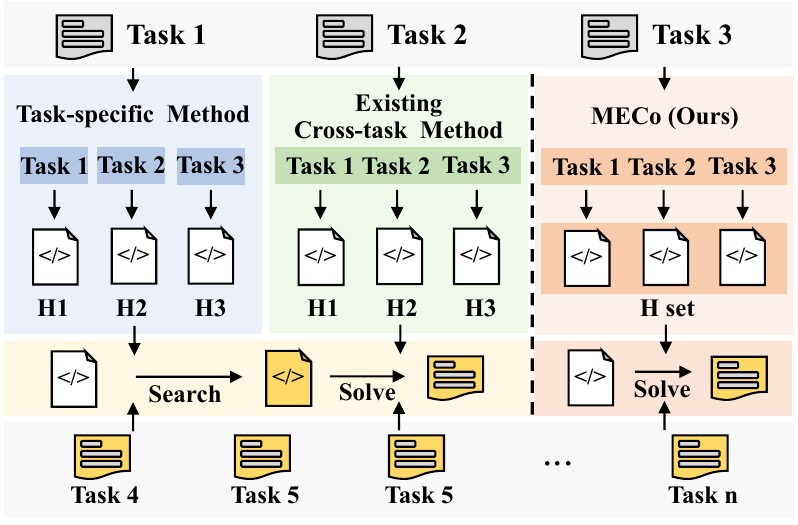}
 \caption{Comparison of heuristic design paradigms. Unlike existing methods, MECo evolves a heuristic set for zero-shot cross-problem generalization.}
 \label{fig:design_settings}
\end{wrapfigure}

Combinatorial optimization problems (COPs) arise in a wide range of logistics and manufacturing applications \citep{toth2014vehicle}. Changes in operational requirements can introduce new problem variants, often requiring existing heuristics to be adapted or redesigned. This process demands domain expertise and repeated cycles of design and evaluation, making heuristic development a recurring cost. Large language models (LLMs) offer a promising way to reduce this burden through automated heuristic design (AHD) \citep{liu2024survey}. Given a problem description and evaluation feedback, these methods iteratively generate and refine executable heuristics. We distinguish task-specific AHD, which designs heuristics for each problem independently, from cross-task AHD, which jointly develops heuristics for multiple problems. \figureautorefname~\ref{fig:design_settings} compares these design paradigms alongside our framework.

We define a task as the design of heuristics for a particular optimization problem. Task-specific AHD methods have made substantial progress in automating this process. Population-based evolutionary search combines LLM-driven program generation with fitness-based selection \citep{liu2024eoh,romeraparedes2024funsearch}. Reflection further guides program generation using performance feedback \citep{ye2024reevo}. Other methods balance exploration and exploitation through diversity-oriented evolution with harmony-search refinement \citep{dat2025hsevo} 
or Monte Carlo tree search \citep{zheng2025mctsahd}. These advances improve heuristic discovery within individual tasks. However, applying the same design framework to another problem typically requires another round of problem-specific search. Without systematic reuse of previously discovered heuristics, even closely related problems can incur repeated design costs.

Cross-task AHD addresses this repetition by jointly developing heuristics for multiple problems. MoH learns reusable heuristic optimizers \citep{shi2026moh}, while MTHS jointly evolves general-purpose metaheuristics and task-specific programs \citep{liu2026mths}. However, jointly designing heuristics for multiple problems does not ensure that the resulting programs can be applied directly to new problems without additional search. In practice, operational requirements can be combined in many ways, and each new combination can define a distinct optimization problem. This creates a potentially large family of related design tasks. If each new task requires a fixed search budget, the total design cost grows with the number of tasks. Moreover, evaluation data for new combinations may be unavailable. This motivates designing heuristics using feedback from a limited set of source tasks and applying them directly to new problems without additional heuristic search.

To address this challenge, we propose MECo, an LLM-driven framework that uses \underline{M}ulti-task \underline{E}volution through task \underline{Co}mbination to enable zero-shot cross-problem generalization. Starting from source tasks, MECo progressively develops new populations for combined task specifications. Their offspring inherit and adapt parent heuristics through evolutionary operators designed for cross-task knowledge transfer. These offspring are evaluated separately on the available datasets, allowing joint feedback to guide heuristic design for combined problems without accessing their instances. To guide interactions among the growing collection of populations, a task transfer gap measures how well the heuristics from one population perform on the sources associated with another. MECo uses this measure to favor compatible parent tasks while exploring new combinations. Finally, it selects a small set from all task populations according to the complementary performance of its members across combinations of sources. Both evolution and set selection rely exclusively on source data, and the selected programs are applied to target problems without further search or adaptation.

Our contributions are summarized as follows.
\begin{itemize}
 \item We introduce MECo, a multi-task evolutionary framework that develops complementary heuristics for zero-shot cross-problem generalization using source task feedback alone.
 \item We introduce a population-level transfer gap to quantify empirical task compatibility and guide heuristic transfer between task populations.
 \item Experiments on 32 variants across vehicle routing (VRP) and flexible job-shop scheduling (FJSP) show lower mean costs on unseen problems than eight AHD baselines under matched proposal budgets. Integrating the framework with three existing AHD methods further improves their performance.
\end{itemize}

\section{Preliminary}
\label{sec:problem}
\subsection{Optimization problems and heuristic-design tasks}

An optimization problem $P$ defines a space of instances $\mathcal X_P$. For each instance $x\in\mathcal X_P$, let $\mathcal Y_P(x)$ be its feasible solution set and $y\in\mathcal Y_P(x)$ a feasible solution. The objective function $F_P(x,y)$ assigns a cost to that solution. We use minimization objectives throughout. Therefore, solving an instance means finding
\begin{equation}
 y_P^\star(x)\in\arg\min_{y\in\mathcal Y_P(x)}F_P(x,y).
 \label{eq:optimization_problem}
\end{equation}

A heuristic-design task $T_P$ seeks a heuristic for solving instances of $P$. Let $\mathcal A$ denote the heuristic space and $D_{T_P}\subset\mathcal X_P$ the evaluation dataset associated with the task. For $h\in\mathcal A$, write $\ell(h,x)$ for the objective cost of its solution on $x$. The single-task AHD objective is to find
\begin{equation}
 h_{T_P}^\star\in\arg\min_{h\in\mathcal A}
 \frac{1}{|D_{T_P}|}\sum_{x\in D_{T_P}}\ell(h,x).
 \label{eq:aad_objective}
\end{equation}
The optimization variable here is the heuristic, which is evaluated across problem instances. In LLM-based AHD, heuristics are represented as programs proposed and refined by an LLM.

\subsection{Source-only design for cross-problem generalization}

We consider source tasks $\mathcal T_{\rm src}$ and target tasks $\mathcal T_{\rm tar}$ whose underlying optimization problems differ with $\mathcal T_{\rm src}\cap\mathcal T_{\rm tar}=\varnothing$. During heuristic design, evaluation feedback is available only from $D_T$ for $T\in\mathcal T_{\rm src}$. Target datasets are kept until testing.

The goal is to obtain a set $\mathcal H\subseteq\mathcal A$ heuristics that performs well across target problems using only the feedback from the source tasks. Within each problem family, the heuristics share a decision representation that permits their application across tasks. Zero-shot evaluation applies the resulting set to target problems without further heuristic search or task-specific adaptation. This setting assesses generalization across out-of-domain problem definitions. Evaluation of new instances of source problems additionally measures the generalization of the in-domain generalization.

\subsection{Compositional problem families}

We instantiate this setting through related problems defined by constraint attributes. Let $P_0$ be a base problem and $\mathcal C=\{c_1,\ldots,c_m\}$ an attribute library. A subset $C\subseteq\mathcal C$ specifies a problem $P_C$, with $P_{\varnothing}=P_0$, and its design task $T_C=T_{P_C}$. Let $T_j=T_{\{c_j\}}$ denote a single-attribute task. The source and target sets are
\begin{equation}
 \begin{aligned}
  \mathcal T_{\rm src} &= \bigl\{T_j\mid 1\leq j\leq m\bigr\}, \\
  \mathcal T_{\rm tar} &= \bigl\{T_C\mid C\subseteq\mathcal C,\ |C|\neq 1\bigr\}.
 \end{aligned}
 \label{eq:task_partition}
\end{equation}
The targets comprise the base problem and combinations of two or more attributes. We call the source tasks roots and write $D_j=D_{T_j}$ when indexing their datasets.

We study VRP and FJSP as two independent problem families under this formulation. Section~\ref{sec:experiment_setting} specifies their attributes and evaluation splits. Appendix~\ref{app:task_definitions} gives the complete problem definitions.

\section{Method}
\label{sec:method}
\subsection{Generalization Through Joint Training}
\label{sec:construction_example}

\begin{wrapfigure}[13]{R}{0.59\textwidth}
 \centering
 \vspace{-18pt}
 \includegraphics[width=\linewidth]{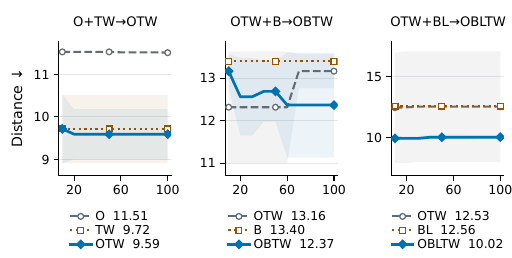}
\caption{A case of generalization to combined VRP variants under separate and joint source training. Curves show the mean cost of the best heuristic in each population.}
 \label{fig:motivation}
\end{wrapfigure}

Joint source training can yield stronger heuristics for a combined problem than separate source training. We illustrate this phenomenon in VRP, where O, TW, B, and L label constraint attributes. Two parent populations evolve on their respective source datasets. An offspring population inherits their heuristics and evolves using both sources, with the same number of additional LLM-generated heuristic proposals for each branch. Figure~\ref{fig:motivation} compares the best heuristic from each branch on the same combined problems and shows that joint source feedback can improve generalization. This observation motivates exchanging heuristics between task populations. 

\subsection{Framework Overview}
\label{sec:framework_overview}

MECo develops heuristics for new task specifications by bringing together populations from existing tasks. For each explored task $T_C$, the population $\mathcal P_C$ contains heuristics intended for $P_C$. The task description guides what these heuristics should solve, and evaluations on its constituent sources guide their selection. Search begins by evolving one population for each source task $T_1,\ldots,T_m$. Each subsequent expansion chooses two existing populations and evolves offspring for their combined task. The resulting population can participate in later expansions.

\begin{figure}[t]
 \centering
 \setlength{\abovecaptionskip}{6pt}
 \includegraphics[width=1\linewidth]{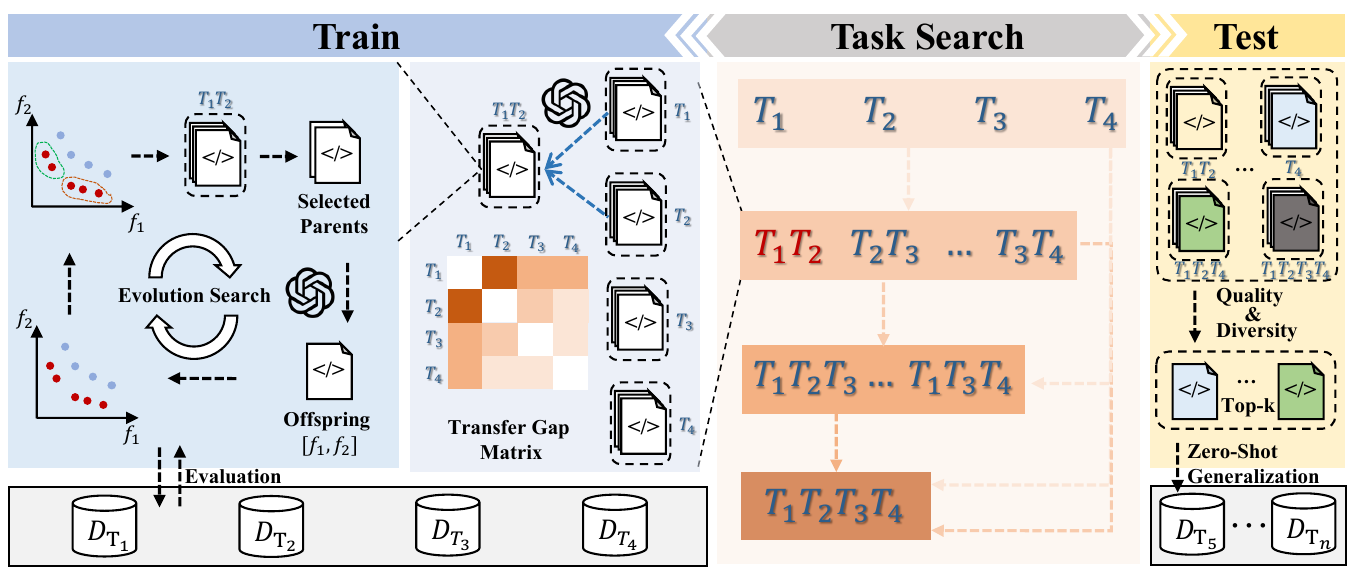}
\caption{MECo couples transfer-guided task interactions with heuristic evolution. Each node represents a task-conditioned population. Source datasets $D_{T_j}$ guide training and final set selection. The selected heuristics are frozen before zero-shot target evaluation.}
 \label{fig:framework}
 \vspace{-6pt}
\end{figure}

Figure~\ref{fig:framework} presents the two search levels. Task-level search determines which populations interact using their measured transfer gaps. Within the chosen task, program-level evolution generates and selects heuristics. The source datasets $D_{T_j}$ remain fixed throughout both levels. After search, MECo selects a small set $\mathcal H$ whose members perform well on different combinations of sources, then freezes it for target evaluation. In the figure, $T_iT_j$ abbreviates the combined task $T_{\{c_i,c_j\}}$.

\subsection{Transfer-guided Evolution}
\label{sec:training}

\paragraph{Source feedback and task quality.}
We evaluate heuristics through normalized source costs and average them over the sources associated with each task,
\begin{equation}
 f_j(h)=\frac{|D_{T_j}|^{-1}\sum_{x\in D_{T_j}}\ell(h,x)}{b_j},
 \qquad
 J_C(h)=\frac{1}{|C|}\sum_{j\,\mid\,c_j\in C}f_j(h),
 \label{eq:local_quality}
\end{equation}
where $f_j(h)$ is the normalized cost on source task $T_j$, $J_C(h)$ measures the quality of $h$ for task $T_C$, and $b_j>0$ is the mean cost of a simple heuristic on $D_{T_j}$, fixed throughout evolution.

The vector $\mathbf f(h)=(f_1(h),\ldots,f_m(h))$ records a heuristic's performance across sources. A source population is evaluated on its own dataset. Offspring for a combined task are evaluated on all sources, with only the constituent entries contributing to $J_C$. For example, an O+TW population uses the average of its normalized O and TW costs for local quality.

\paragraph{Choosing interacting tasks.}
\label{sec:search_structure}
\label{sec:transfer}
Given these source scores, MECo chooses two populations to develop a combined task. A pair $\mathcal P_A,\mathcal P_B$ is eligible when $C=A\cup B$ contains attributes contributed by both parents, so $A\subsetneq C$ and $B\subsetneq C$. Thus a combined task such as O+TW can later interact with L to develop O+TW+L. Different parent pairs can reach the same task, and their populations are merged after its inner evolution search.

Among eligible pairs, transfer gap measures the change in average population quality on the destination task. We define the transferred and native scores as
\[
 s(A\!\rightarrow\!B)=\frac{1}{|\mathcal P_A|}\sum_{h\in\mathcal P_A}J_B(h),
 \qquad
 s(B\!\rightarrow\!B)=\frac{1}{|\mathcal P_B|}\sum_{h\in\mathcal P_B}J_B(h).
\]
Both scores use the complete datasets of the sources associated with $T_B$. The directional and symmetric transfer gaps are
\begin{align}
 g(A\!\rightarrow\!B)
 &=\max\left\{0,\,
 \frac{s(A\!\rightarrow\!B)-s(B\!\rightarrow\!B)}
 {\max\{|s(B\!\rightarrow\!B)|,\epsilon\}}
 \right\},
 \label{eq:directional_transfer_gap}\\
 G(A,B)
 &=\frac{g(A\!\rightarrow\!B)+g(B\!\rightarrow\!A)}{2}.
 \label{eq:transfer_gap}
\end{align}
$G(A,B)$ averages the relative transfer losses in both directions.

MECo first prioritizes pairs leading to unexplored combined tasks and then favors a smaller $G(A,B)$. Each task pair is retired after a successful expansion, while either task remains available to interact with other tasks. Different pairs can therefore contribute to the same combined task.

\paragraph{Evolving Task-Conditioned Heuristics.}
\label{sec:generation}
Once the parent tasks are chosen, MECo uses their best heuristics as the initial parents for $T_C$. The LLM receives the parent program or programs required by its operator, their task descriptions and source scores, and the combined-task description. Later generations choose parents from the local population and evaluate each proposed heuristic on the source datasets.

Four prompt-based operators guide this evolution through three forms of crossover and one guided mutation: \textbf{(1) $\mathrm{c}_1$} performs mechanism composition, combining compatible decision rules from two parents. \textbf{(2) $\mathrm{c}_2$} performs conflict-resolving crossover, integrating conflicting parent rules through conditional or hierarchical decisions. \textbf{(3) $\mathrm{c}_3$} performs exploratory crossover, using two parents with contrasting source scores to propose a distinct decision principle for an under-covered source dimension. \textbf{(4) $\mathrm{m}_1$} performs transfer-oriented mutation, retaining the parent's backbone while modifying a decision mechanism to improve its source performance.

The local population preserves candidates with different source strengths. Candidates are ranked by nondominated fronts on the constituent dimensions of $\mathbf f(h)$, with crowding distance favoring less crowded score vectors within each front \citep{deb2002nsgaii}. Comparing their decisions on the same construction states further prioritizes distinct behaviors \citep{zhang2026rethinking}.

After the evolutionary search, successful heuristics from that expansion are selected by $J_C$ and behavioral priority to form the retained population. If $T_C$ already exists, its previous and new populations are merged using the local survival rule. The updated population then becomes available for future task interactions. Repeating these expansions produces candidate heuristics for progressively richer task specifications.

\subsection{Heuristic Set Selection for Zero-Shot Evaluation}
\label{sec:outputs}
After evolution, we collect representatives from all task populations, complete their source evaluations, and filter repeated behaviors to obtain a candidate pool $\mathcal E$. Following the principle of complementary heuristic selection \citep{xu2010hydra}, we choose $K$ members to cover combinations of sources. To direct selection toward generalization beyond the source tasks, the coverage objective ranges over combined task specifications $\mathcal C_{\geq2}=\{C\subseteq\mathcal C\mid |C|\geq2\}$. Candidates from source populations remain eligible. For a nonempty set $\mathcal H$, define

\begin{equation}
 J_{\rm comp}(\mathcal H)=
 \frac{1}{|\mathcal C_{\geq2}|}
 \sum_{C\in\mathcal C_{\geq2}}\min_{h\in\mathcal H}J_C(h).
 \label{eq:portfolio_proxy}
\end{equation}
For each combination $C$, we first compute every member's mean cost on the constituent sources, $J_C(h)$, and retain the lowest value in the set. We then average these values across combinations. The objective rewards their complementary strengths.

We begin with the heuristic of lowest mean normalized source cost. Each additional member is chosen by how much it reduces the current set score,
\begin{equation}
 h^\star\in\arg\max_{h\in\mathcal E\setminus\mathcal H}
 \left[J_{\rm comp}(\mathcal H)-
 J_{\rm comp}(\mathcal H\cup\{h\})\right]
 \label{eq:marginal_gain}
\end{equation}
until $|\mathcal H|=K$. Ties in marginal gain favor the candidate with lower mean normalized source cost. We use $K=4$, freeze the selected programs and set membership, and evaluate all members on each target problem without further search or adaptation.


\label{sec:method_end}

\section{Experiments}
\label{sec:experiments}
\subsection{Experimental setup}
\label{sec:experiment_setting}

\paragraph{Problem setup.}
We study vehicle routing problem (VRP) and flexible job-shop scheduling problem (FJSP). The base VRP is the capacitated vehicle routing problem (CVRP), which serves every customer using depot-to-depot routes subject to vehicle capacity and minimizes total travel distance. The base FJSP assigns each operation to an eligible machine and sequences operations to minimize makespan, subject to within-job precedence and machine availability. For VRP, the designed heuristic selects the next customer during route construction. For FJSP, it selects the next operation and its machine from feasible assignments. 

We construct 16 variants per family by activating subsets of 4 optional constraints. VRP uses open routes (O), backhauls (B), time windows (TW), and route-length limits (L), following multi-task VRP formulations \citep{liu2024multitaskrouting}. FJSP uses release dates (R), assembly precedence (A), sequence-dependent setups (S), and transport lags (T) \citep{reijnen2026benchmark,lin2022assembly,shen2018setup,berterottiere2024transportation}. Only the four single-constraint tasks provide training datasets. Testing on larger, unseen instances of these four problems measures in-domain (ID) generalization. Testing on the 11 combined-constraint problems and the base problem measures out-of-domain (OOD) generalization. Appendix~\ref{app:task_definitions} specifies the constraint semantics.

\paragraph{Baselines.}
We compare eight AHD methods and two classical construction rules per family, organized into three groups.

\noindent\textbf{(1) Task-specific AHD.} EoH \citep{liu2024eoh}, ReEvo \citep{ye2024reevo}, MCTS-AHD \citep{zheng2025mctsahd}, EoH-S \citep{liu2025eohs}, MEoH \citep{yao2025meoh}, and HSEvo \citep{dat2025hsevo} perform heuristic search for a given design task. 

\noindent\textbf{(2) Cross-task AHD.} MoH co-evolves heuristics and optimizers \citep{shi2026moh}, while EMO-STA combines shared search with task-specific adaptation \citep{gozeten2026emo}. All methods use the same four source tasks and problem interfaces.

\noindent\textbf{(3) Classical heuristics.} VRP uses nearest-feasible construction and savings-priority construction \citep{clarke1964scheduling}. FJSP uses earliest-completion-time (ECT) dispatch and most-work-remaining dispatch with ECT machine assignment (MWKR-ECT), following established dispatching principles \citep{lei2022multiaction}. Appendix~\ref{app:baseline_parameters} gives the baseline settings and rule definitions.

\begin{wraptable}[8]{r}{0.42\textwidth}
 \centering\normalsize
 \vspace{-10pt}
 \caption{Data per family. Instances are counted per variant.}
 \label{tab:data_setting}
 \resizebox{\linewidth}{!}{%
 \begin{tabular}{lrrr}
 \toprule
 Split & Variants & Instances & Size \\
 \midrule
 Training & 4 & 10 & 50 \\
 ID test & 4 & 1,000 & 100 \\
 OOD test & 12 & 1,000 & 100 \\
 \bottomrule
 \end{tabular}%
 }
\end{wraptable}

\paragraph{Search and data settings.}
All AHD searches use GPT-4o-mini with a budget of 2,000 new algorithms per family and select 4 best heuristics for final testing. Following feasibility-masked construction \citep{liu2024multitaskrouting}, common task-specific masks enforce constraints for all methods. \tableautorefname~\ref{tab:data_setting} summarizes the data. Size counts customers for VRP and operations for FJSP. Computation uses Intel Xeon Gold 6348 CPUs at 2.60\,GHz. Appendix~\ref{app:problem_parameters} details data construction.
\par\WFclear

\paragraph{Evaluation.}
For the test dataset $D_{T_C}^{\mathrm{test}}$ of task $T_C$, we report
\begin{equation}
 \operatorname{Best4}(C)=\min_{h\in\mathcal H}\frac{1}{|D_{T_C}^{\mathrm{test}}|}\sum_{x\in D_{T_C}^{\mathrm{test}}}\ell(h,x).
 \label{eq:test_metrics}
\end{equation}
Best4 is the lowest whole-dataset mean cost among the four selected heuristics. Unless stated otherwise, we repeat each AHD experiment three times and report mean and sample standard deviation of the resulting Best4 scores, averaged over all 16, 4 ID, or 12 OOD variants. Deterministic classical rules are evaluated once. Relative gap is the percentage excess over the best observed method mean on the same variant. Gap and Rank average these gaps and per-variant ranks over all 16 variants. Bold and underlined means indicate first and second place, including ties. Green rows highlight the corresponding best methods, and $\#k$ denotes the number of variants.

\subsection{Cross-problem generalization}
\label{sec:main_results}

\begin{table}[t]
 \centering\normalsize
 \caption{Performance comparison across the two problem families.}
 \label{tab:main}
 \ifdefined\MECoMeanWidth\else\newlength{\MECoMeanWidth}\fi
\begingroup
\settowidth{\MECoMeanWidth}{$\mathbf{00.00}$}
\def\MECoMainRows{
\toprule
\multicolumn{6}{@{}l@{}}{\textbf{(a) VRP distance} $\downarrow$} \\
Method & All (\#16) & ID (\#4) & OOD (\#12) & \shortstack{Rank\\(\#16)} & \shortstack{Gap (\%)\\(\#16)} \\
\midrule
Nearest-feasible & \MECoStat{00.00}{0.00}{23.53}{} & \MECoStat{00.00}{0.00}{23.04}{} & \MECoStat{00.00}{0.00}{23.69}{} & $7.66$ & $7.74$ \\
Savings-priority & \MECoStat{00.00}{0.00}{24.51}{} & \MECoStat{00.00}{0.00}{22.48}{} & \MECoStat{00.00}{0.00}{25.19}{} & $7.75$ & $14.64$ \\
\midrule
EoH & \MECoStat{00.00}{0.00}{23.16}{0.80} & \MECoStat{00.00}{0.00}{22.14}{0.08} & \MECoStat{00.00}{0.00}{23.49}{1.04} & $7.12$ & $7.52$ \\
ReEvo & \MECoStat{00.00}{0.00}{23.20}{0.55} & \MECoStat{00.00}{0.00}{22.41}{0.46} & \MECoStat{00.00}{0.00}{23.46}{0.60} & $8.78$ & $7.58$ \\
MCTS-AHD & \MECoStat{00.00}{0.00}{22.54}{0.17} & \MECoStat{00.00}{0.00}{22.10}{0.05} & \MECoStat{00.00}{0.00}{22.69}{0.21} & $5.81$ & $4.76$ \\
EoH-S & \MECoStat{00.00}{0.00}{22.75}{0.10} & \MECoStat{00.00}{0.00}{22.18}{0.12} & \MECoStat{00.00}{0.00}{22.94}{0.12} & $6.94$ & $5.64$ \\
MEoH & \MECoStat{00.00}{0.00}{22.94}{0.03} & \MECoStat{00.00}{0.00}{22.53}{0.19} & \MECoStat{00.00}{0.00}{23.08}{0.08} & $6.22$ & $5.83$ \\
HSEvo & \MECoStat{00.00}{0.00}{22.47}{0.27} & \MECoStat{00.00}{0.00}{22.12}{0.11} & \MECoStat{00.00}{0.00}{22.58}{0.33} & $\underline{3.97}$ & $4.34$ \\
\midrule
EMO-STA & \MECoStat{00.00}{0.00}{23.01}{0.50} & \MECoStat{00.00}{0.00}{22.46}{0.68} & \MECoStat{00.00}{0.00}{23.20}{0.45} & $5.28$ & $5.02$ \\
MoH & \MECoStat{00.00}{0.00}{\underline{22.39}}{0.62} & \MECoStat{00.00}{0.00}{\underline{21.88}}{0.80} & \MECoStat{00.00}{0.00}{\underline{22.55}}{0.57} & $4.16$ & $\underline{2.82}$ \\
\midrule
\rowcolor{BestGreen}\textbf{MECo} & \MECoStat{00.00}{0.00}{\mathbf{21.98}}{0.50} & \MECoStat{00.00}{0.00}{\mathbf{21.60}}{0.70} & \MECoStat{00.00}{0.00}{\mathbf{22.11}}{0.43} & $\mathbf{2.31}$ & $\mathbf{1.40}$ \\
\bottomrule
}
\begin{tabularx}{\linewidth}{lYYYrr}\MECoMainRows\end{tabularx}
\endgroup
\par\medskip
\begingroup
\settowidth{\MECoMeanWidth}{$\mathbf{0000.0}$}
\def\MECoMainRows{
\toprule
\multicolumn{6}{@{}l@{}}{\textbf{(b) FJSP makespan} $\downarrow$} \\
Method & All (\#16) & ID (\#4) & OOD (\#12) & \shortstack{Rank\\(\#16)} & \shortstack{Gap (\%)\\(\#16)} \\
\midrule
ECT & \MECoStat{0000.0}{00.0}{1486.8}{} & \MECoStat{0000.0}{00.0}{1301.4}{} & \MECoStat{0000.0}{00.0}{1548.6}{} & $8.44$ & $5.88$ \\
MWKR-ECT & \MECoStat{0000.0}{00.0}{1751.7}{} & \MECoStat{0000.0}{00.0}{1436.3}{} & \MECoStat{0000.0}{00.0}{1856.9}{} & $11.00$ & $23.64$ \\
\midrule
EoH & \MECoStat{0000.0}{00.0}{1466.8}{33.5} & \MECoStat{0000.0}{00.0}{1247.0}{37.6} & \MECoStat{0000.0}{00.0}{1540.1}{32.2} & $7.50$ & $4.00$ \\
ReEvo & \MECoStat{0000.0}{00.0}{1446.0}{29.7} & \MECoStat{0000.0}{00.0}{1229.2}{12.0} & \MECoStat{0000.0}{00.0}{1518.2}{35.8} & $5.25$ & $2.54$ \\
MCTS-AHD & \MECoStat{0000.0}{00.0}{1456.8}{47.4} & \MECoStat{0000.0}{00.0}{1239.7}{47.9} & \MECoStat{0000.0}{00.0}{1529.2}{48.1} & $6.56$ & $3.35$ \\
EoH-S & \MECoStat{0000.0}{00.0}{1460.2}{45.7} & \MECoStat{0000.0}{00.0}{1241.7}{45.3} & \MECoStat{0000.0}{00.0}{1533.1}{47.3} & $6.91$ & $3.58$ \\
MEoH & \MECoStat{0000.0}{00.0}{1453.0}{36.7} & \MECoStat{0000.0}{00.0}{1237.2}{26.2} & \MECoStat{0000.0}{00.0}{1525.0}{42.1} & $6.00$ & $3.06$ \\
HSEvo & \MECoStat{0000.0}{00.0}{1443.3}{23.3} & \MECoStat{0000.0}{00.0}{1225.2}{5.1} & \MECoStat{0000.0}{00.0}{1516.0}{29.5} & $4.81$ & $2.32$ \\
\midrule
EMO-STA & \MECoStat{0000.0}{00.0}{1447.6}{32.4} & \MECoStat{0000.0}{00.0}{1255.7}{38.2} & \MECoStat{0000.0}{00.0}{1511.5}{31.1} & $5.81$ & $2.93$ \\
MoH & \MECoStat{0000.0}{00.0}{\underline{1421.4}}{12.2} & \MECoStat{0000.0}{00.0}{\underline{1223.5}}{13.0} & \MECoStat{0000.0}{00.0}{\underline{1487.3}}{11.9} & $\underline{2.59}$ & $\underline{0.95}$ \\
\midrule
\rowcolor{BestGreen}\textbf{MECo} & \MECoStat{0000.0}{00.0}{\mathbf{1408.7}}{1.1} & \MECoStat{0000.0}{00.0}{\mathbf{1211.3}}{3.5} & \MECoStat{0000.0}{00.0}{\mathbf{1474.4}}{1.1} & $\mathbf{1.12}$ & $\mathbf{0.02}$ \\
\bottomrule
}
\begin{tabularx}{\linewidth}{lYYYrr}\MECoMainRows\end{tabularx}
\endgroup

\end{table}

As shown in \tableautorefname~\ref{tab:main}, MECo achieves the lowest mean costs on both the source and unseen problems in each family. Its advantage extends across the all-variant, ID, and OOD aggregates, together with the lowest average rank and relative gap. Under the shared source-only protocol, MECo therefore achieves better aggregate performance on unseen problems than the compared AHD methods.

Specifically, on VRP, MoH is the strongest baseline by mean cost in all three groups. MECo lowers its ID cost from $21.88$ to $21.60$ and its OOD cost from $22.55$ to $22.11$. Therefore, the improvement extends beyond the four problems supplying training feedback. In addition, MECo also has an average rank of $2.31$, compared with $3.97$ for the next-best method by rank, HSEvo. Similarly, on FJSP, MECo lowers the OOD mean from $1487.3$ for MoH to $1474.4$, alongside a lower ID mean of $1211.3$. Its OOD standard deviation is also smaller, $1.1$ compared with $11.9$ for MoH, indicating less variation across the repeated searches in this study. The average rank of $1.12$ and mean relative gap of $0.02\%$ place MECo close to the best observed method across the 16 variants.

\begin{figure}[!htbp]
 \centering
 \includegraphics[width=0.90\linewidth]{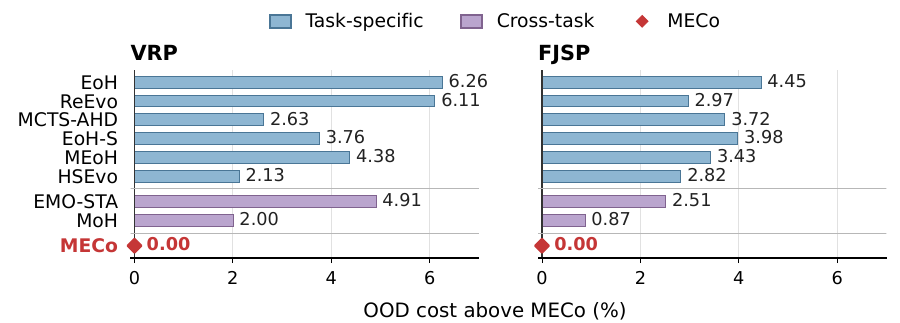}
 \caption{OOD mean cost relative to MECo, computed as $100(\bar f/\bar f_{\mathrm{MECo}}-1)$.}
 \label{fig:combination_coverage}
\end{figure}

As shown in \figureautorefname~\ref{fig:combination_coverage}, both task-specific and cross-task baselines have higher OOD aggregate costs in both families. Together, these observations support the advantage of MECo on cross-problem generalization across VRP and FJSP. Appendix~\ref{app:all_variant_results} provides more detailed results.

\subsection{Framework effectiveness}
\label{sec:engine_adaptation}

We further investigate the effectiveness of MECo's task-search framework across different heuristic search methods. For each baseline $X$, we construct $X$-MECo by using its native heuristic generation and population update rules within MECo's task populations. MECo chooses the interacting tasks and transfers their parent heuristics to initialize a combined-task population. The baseline then evolves that population using its own search rules. Thus, EoH-MECo runs EoH within each task population while MECo determines which tasks interact. ReEvo-MECo and HSEvo-MECo follow the same construction. After search, MECo selects the final four heuristics from the task populations.

\begin{table}[!htbp]
 \centering\normalsize
 \caption{Baseline engines with and without MECo.}
 \label{tab:engine_transfer}
 
\resizebox{\linewidth}{!}{%
\begin{tabular}{lrrrrrr}
\toprule
 & \multicolumn{3}{c}{VRP distance $\downarrow$} & \multicolumn{3}{c}{FJSP makespan $\downarrow$} \\
\cmidrule(lr){2-4}\cmidrule(lr){5-7}
Method & All (\#16) & ID (\#4) & OOD (\#12) & All (\#16) & ID (\#4) & OOD (\#12) \\
\midrule
EoH & \MECoStat{00.00}{0.00}{23.16}{0.80} & \MECoStat{00.00}{0.00}{22.14}{0.08} & \MECoStat{00.00}{0.00}{23.49}{1.04} & \MECoStat{0000.0}{00.0}{1466.8}{33.5} & \MECoStat{0000.0}{00.0}{1247.0}{37.6} & \MECoStat{0000.0}{00.0}{1540.1}{32.2} \\
\rowcolor{BestGreen}EoH-MECo & \MECoStat{00.00}{0.00}{\mathbf{21.99}}{0.66} & \MECoStat{00.00}{0.00}{\mathbf{21.66}}{0.83} & \MECoStat{00.00}{0.00}{\mathbf{22.10}}{0.61} & \MECoStat{0000.0}{00.0}{\mathbf{1432.0}}{6.5} & \MECoStat{0000.0}{00.0}{\mathbf{1211.1}}{13.3} & \MECoStat{0000.0}{00.0}{\mathbf{1505.7}}{4.9} \\
\midrule
ReEvo & \MECoStat{00.00}{0.00}{23.20}{0.55} & \MECoStat{00.00}{0.00}{22.41}{0.46} & \MECoStat{00.00}{0.00}{23.46}{0.60} & \MECoStat{0000.0}{00.0}{1446.0}{29.7} & \MECoStat{0000.0}{00.0}{1229.2}{12.0} & \MECoStat{0000.0}{00.0}{1518.2}{35.8} \\
\rowcolor{BestGreen}ReEvo-MECo & \MECoStat{00.00}{0.00}{\mathbf{22.86}}{1.75} & \MECoStat{00.00}{0.00}{\mathbf{21.17}}{0.70} & \MECoStat{00.00}{0.00}{\mathbf{23.43}}{2.41} & \MECoStat{0000.0}{00.0}{\mathbf{1415.9}}{28.4} & \MECoStat{0000.0}{00.0}{\mathbf{1207.3}}{18.0} & \MECoStat{0000.0}{00.0}{\mathbf{1485.4}}{32.2} \\
\midrule
HSEvo & \MECoStat{00.00}{0.00}{22.47}{0.27} & \MECoStat{00.00}{0.00}{22.12}{0.11} & \MECoStat{00.00}{0.00}{22.58}{0.33} & \MECoStat{0000.0}{00.0}{1443.3}{23.3} & \MECoStat{0000.0}{00.0}{1225.2}{5.1} & \MECoStat{0000.0}{00.0}{1516.0}{29.5} \\
\rowcolor{BestGreen}HSEvo-MECo & \MECoStat{00.00}{0.00}{\mathbf{22.14}}{0.13} & \MECoStat{00.00}{0.00}{\mathbf{21.83}}{0.09} & \MECoStat{00.00}{0.00}{\mathbf{22.24}}{0.19} & \MECoStat{0000.0}{00.0}{\mathbf{1418.1}}{17.5} & \MECoStat{0000.0}{00.0}{\mathbf{1218.5}}{16.4} & \MECoStat{0000.0}{00.0}{\mathbf{1484.6}}{17.8} \\
\bottomrule
\end{tabular}%
}

\end{table}

As shown in \tableautorefname~\ref{tab:engine_transfer}, all three hybrids achieve lower mean costs than their baselines on both ID and OOD variants in both families. In particular, EoH-MECo reduces mean OOD cost relative to EoH over $5\%$ on VRP and $2\%$ on FJSP. These results support the effectiveness of the framework beyond MECo's population evolution rules.

\noindent\begin{minipage}{\linewidth}
\begin{wrapfigure}{r}{0.55\textwidth}
 \centering
 \vspace{-15pt}
 \setlength{\abovecaptionskip}{14pt}
 \includegraphics[width=0.90\linewidth]{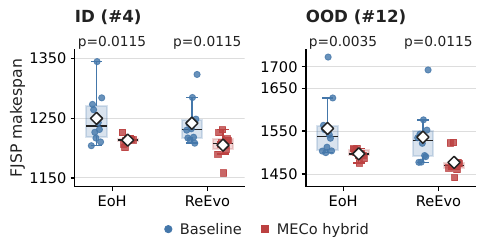}
 \caption{Performance on FJSP for ten runs.}
 \label{fig:fjsp_engine_significance}
\end{wrapfigure}

To assess reliability, we extend the FJSP comparison to ten independent runs per method, comparing EoH and ReEvo with their MECo counterparts. As shown in \figureautorefname \ref{fig:fjsp_engine_significance}, both MECo variants achieve lower mean costs on ID and OOD tasks. All four improvements are statistically significant based on two-sided independent-samples permutation tests with Holm correction across the four comparisons. Appendix~\ref{app:engine_transfer} provides the experimental and statistical details.
\par\WFclear

\end{minipage}

\subsection{Component ablations}
\label{sec:main_ablations}

We study the importance of task pairing and heuristic evolution operators in MECo by evaluating five ablation settings against the complete framework. The condition without transfer gap replaces guided task-pair selection with uniform random sampling, while the remaining four conditions individually omit mechanism composition, conflict resolution, exploratory crossover, or transfer-oriented mutation across all evolutionary stages. As shown in \tableautorefname~\ref{tab:search_ablation}, removing any single component degrades overall solution quality, leading to higher mean costs across both problem families on both ID and OOD tasks. Specifically, omitting transfer gap guidance and transfer-oriented mutation leads to the largest performance drops on VRP. On FJSP, removing exploratory crossover yields the highest makespan while also introducing substantial run-to-run variance. In addition, full MECo consistently maintains lower standard deviations than the ablated variants, indicating that the full pipeline provides more reliable optimization. Together, these results confirm that guided task pairing and the specialized operators work together to achieve MECo's strong generalization across diverse problem domains.   

\begin{table}[!htbp]
 \centering\normalsize
 \caption{Component ablations with full MECo.}
 \label{tab:search_ablation}
 
\resizebox{\linewidth}{!}{%
\begin{tabular}{lrrrrrr}
\toprule
 & \multicolumn{3}{c}{VRP distance $\downarrow$} & \multicolumn{3}{c}{FJSP makespan $\downarrow$} \\
\cmidrule(lr){2-4}\cmidrule(lr){5-7}
Condition & All (\#16) & ID (\#4) & OOD (\#12) & All (\#16) & ID (\#4) & OOD (\#12) \\
\midrule
w/o transfer gap & \MECoStat{00.00}{0.00}{22.35}{0.05} & \MECoStat{00.00}{0.00}{22.16}{0.05} & \MECoStat{00.00}{0.00}{22.41}{0.05} & \MECoStat{0000.0}{00.0}{\underline{1410.2}}{7.8} & \MECoStat{0000.0}{00.0}{\underline{1214.6}}{11.5} & \MECoStat{0000.0}{00.0}{\underline{1475.3}}{6.6} \\
w/o $\mathrm{c}_1$ & \MECoStat{00.00}{0.00}{22.28}{0.16} & \MECoStat{00.00}{0.00}{22.11}{0.11} & \MECoStat{00.00}{0.00}{\underline{22.34}}{0.17} & \MECoStat{0000.0}{00.0}{1418.4}{11.3} & \MECoStat{0000.0}{00.0}{1226.7}{13.9} & \MECoStat{0000.0}{00.0}{1482.3}{10.7} \\
w/o $\mathrm{c}_2$ & \MECoStat{00.00}{0.00}{\underline{22.19}}{0.09} & \MECoStat{00.00}{0.00}{\underline{21.73}}{0.31} & \MECoStat{00.00}{0.00}{\underline{22.34}}{0.06} & \MECoStat{0000.0}{00.0}{1411.9}{8.7} & \MECoStat{0000.0}{00.0}{1216.7}{13.2} & \MECoStat{0000.0}{00.0}{1477.0}{7.3} \\
w/o $\mathrm{c}_3$ & \MECoStat{00.00}{0.00}{22.32}{0.11} & \MECoStat{00.00}{0.00}{21.89}{0.31} & \MECoStat{00.00}{0.00}{22.46}{0.16} & \MECoStat{0000.0}{00.0}{1424.0}{24.7} & \MECoStat{0000.0}{00.0}{1224.0}{18.5} & \MECoStat{0000.0}{00.0}{1490.7}{26.7} \\
w/o $\mathrm{m}_1$ & \MECoStat{00.00}{0.00}{22.37}{0.09} & \MECoStat{00.00}{0.00}{22.08}{0.05} & \MECoStat{00.00}{0.00}{22.47}{0.10} & \MECoStat{0000.0}{00.0}{1411.4}{4.9} & \MECoStat{0000.0}{00.0}{1215.2}{3.9} & \MECoStat{0000.0}{00.0}{1476.8}{5.2} \\
\midrule
\rowcolor{BestGreen}\textbf{MECo} & \MECoStat{00.00}{0.00}{\mathbf{21.98}}{0.50} & \MECoStat{00.00}{0.00}{\mathbf{21.60}}{0.70} & \MECoStat{00.00}{0.00}{\mathbf{22.11}}{0.43} & \MECoStat{0000.0}{00.0}{\mathbf{1408.7}}{1.1} & \MECoStat{0000.0}{00.0}{\mathbf{1211.3}}{3.5} & \MECoStat{0000.0}{00.0}{\mathbf{1474.4}}{1.1} \\
\bottomrule
\end{tabular}%
}

\end{table}

\subsection{Generation-model sensitivity}
\label{sec:model_ablation}

To examine how the underlying generator model affects MECo, we compare gpt-4o-mini, gpt-5.4-nano, and gemini-3.1-flash-lite under the same budget of 2,000 heuristic proposals. As shown in \tableautorefname~\ref{tab:model_ablation}, gemini-3.1-flash-lite achieves the best results across all settings, yielding the lowest costs on both ID and OOD tasks for both VRP and FJSP. Meanwhile, the relative performance between the other two models depends on the problem type. gpt-4o-mini outperforms gpt-5.4-nano on VRP across all splits, but this trend reverses on FJSP, where gpt-5.4-nano consistently achieves shorter makespans. This indicates that while MECo benefits from stronger generator models, smaller LLMs still show distinct domain preferences between VRP and FJSP tasks.

\begin{table}[!htbp]
 \centering\normalsize
 \caption{MECo performance with different generation models.}
 \label{tab:model_ablation}
 
\resizebox{\linewidth}{!}{%
\begin{tabular}{lrrrrrr}
\toprule
 & \multicolumn{3}{c}{VRP distance $\downarrow$} & \multicolumn{3}{c}{FJSP makespan $\downarrow$} \\
\cmidrule(lr){2-4}\cmidrule(lr){5-7}
Model & All (\#16) & ID (\#4) & OOD (\#12) & All (\#16) & ID (\#4) & OOD (\#12) \\
\midrule
gpt-4o-mini & \MECoStat{00.00}{0.00}{\underline{21.98}}{0.50} & \MECoStat{00.00}{0.00}{\underline{21.60}}{0.70} & \MECoStat{00.00}{0.00}{\underline{22.11}}{0.43} & \MECoStat{0000.0}{00.0}{1408.7}{1.1} & \MECoStat{0000.0}{00.0}{1211.3}{3.5} & \MECoStat{0000.0}{00.0}{1474.4}{1.1} \\
gpt-5.4-nano & \MECoStat{00.00}{0.00}{22.29}{0.46} & \MECoStat{00.00}{0.00}{21.72}{0.43} & \MECoStat{00.00}{0.00}{22.48}{0.48} & \MECoStat{0000.0}{00.0}{\underline{1405.8}}{1.7} & \MECoStat{0000.0}{00.0}{\underline{1205.1}}{3.4} & \MECoStat{0000.0}{00.0}{\underline{1472.7}}{1.9} \\
\rowcolor{BestGreen}gemini-3.1-flash-lite & \MECoStat{00.00}{0.00}{\mathbf{21.72}}{0.36} & \MECoStat{00.00}{0.00}{\mathbf{21.36}}{0.28} & \MECoStat{00.00}{0.00}{\mathbf{21.84}}{0.38} & \MECoStat{0000.0}{00.0}{\mathbf{1367.6}}{22.3} & \MECoStat{0000.0}{00.0}{\mathbf{1163.2}}{13.6} & \MECoStat{0000.0}{00.0}{\mathbf{1435.8}}{25.1}\\
\bottomrule
\end{tabular}%
}

\end{table}

\paragraph{Further analysis.}
We study the adaptation on VRP variants OB and BLTW. MECo shares 500 additional proposals across both targets, while the task-specific baselines use 500 per target. Although MECo is not the best on either target individually, its costs remain close to the best observed values and its average rank is the lowest, using half the total proposal allowance. Appendix~\ref{app:routing_adaptation} reports these results. We further study the joint design across TSP, VRP, and FJSP under a shared budget of 1,000 proposals. MECo yields TSP and VRP costs that are lower than those of EMO-STA and MoH at both test sizes. It also improves over EMO-STA on FJSP, where MoH performs best. This extends joint design beyond variants of one problem family, as reported in Appendix~\ref{app:heterogeneous_tasks}. Appendix~\ref{app:resource_accounting} reports resource use.

\section{Conclusion}
\label{sec:conclusion}
This work introduces MECo, an LLM-driven multi-task evolutionary framework for zero-shot cross-problem generalization. Transfer gaps guide interactions among task populations, enabling heuristics to be adapted and recombined. Source feedback supports both evolution and complementary set selection, without target evaluations during design. Experiments on VRP and FJSP demonstrate improved ID and OOD performance under a shared design budget. Moreover, integrating MECo with EoH, ReEvo, and HSEvo consistently improves their performance. Together, these results support coordinated multi-task search as an effective approach to designing heuristic sets that generalize beyond their source problems.
\label{sec:main_end}

\subsection*{AI use statement}
\phantomsection
\label{sec:ai_use}

In this work, we used generative AI tools for assisting with translation. We have not used generative AI tools for generating synthetic data sets, assisting in the writing of proofs, proposing or refining hypotheses, designing or providing feedback on research methodology or experiments, implementing methods, cleaning and reformatting datasets, supporting qualitative and thematic data analysis, interpreting results, helping develop theoretical models or conceptual frameworks, formulating mathematical claims, or providing critical ingredients for proving mathematical claims, as these uses are not applicable to this work. Additionally, we used generative AI tools for editing a research paper to improve readability. We have reviewed all AI-assisted work. LLM-generated translation was verified for correctness by 2 authors. We take responsibility for the final content of this work, including text, claims, or artifacts produced with the aid of generative AI.

\bibliography{iclr2027_conference}
\bibliographystyle{iclr2027_conference}

\clearpage
\appendix
\section*{Appendix contents}
\label{app:contents}

\begingroup
\normalsize
\setlength{\parindent}{0pt}
\setlength{\parskip}{0pt}
\setlength{\baselineskip}{14pt}
\newcommand{\appendixchapterentry}[2]{%
 \par\addvspace{5.5pt}\noindent
 \hyperref[#1]{\makebox[\linewidth][l]{%
  \bfseries\makebox[2.7em][l]{\ref*{#1}}#2\hspace{0.7em}%
  \dotfill\hspace{0.7em}\makebox[2em][r]{\pageref*{#1}}}}\par}
\newcommand{\appendixsubentry}[2]{%
 \noindent\hyperref[#1]{\makebox[\linewidth][l]{%
  \hspace*{1.5em}\makebox[3.2em][l]{\ref*{#1}}#2\hspace{0.7em}%
  \dotfill\hspace{0.7em}\makebox[2em][r]{\pageref*{#1}}}}\par}

\appendixchapterentry{sec:related_work}{Related work}
\appendixsubentry{app:related_task_specific}{Task-specific automated heuristic design}
\appendixsubentry{app:related_cross_task}{Cross-task automated heuristic design}
\appendixsubentry{app:related_transfer}{Knowledge transfer across optimization tasks}

\appendixchapterentry{app:implementation}{More Method Details}
\appendixsubentry{app:search_allocation}{Search Procedure}
\appendixsubentry{app:transfer_details}{Source feedback and transfer}
\appendixsubentry{app:parent_programs}{Heuristic Generation}
\appendixsubentry{app:survival_behavior}{Population survival}
\appendixsubentry{app:selection}{Final heuristic-set selection}
\appendixsubentry{app:proposal_accounting}{Construction and candidate validity}

\appendixchapterentry{app:task_definitions}{Task definitions and constraint semantics}
\appendixsubentry{app:vrp_constraints}{VRP}
\appendixsubentry{app:fjsp_constraints}{FJSP}

\appendixchapterentry{app:benchmark_parameters}{Benchmark variants and experimental parameters}
\appendixsubentry{app:variant_specification}{Complete variant specification}
\appendixsubentry{app:problem_parameters}{Problem parameters and data construction}
\appendixsubentry{app:baseline_parameters}{Search and baseline parameters}
\appendixsubentry{app:classical_rules}{Classical constructive reference rules}

\appendixchapterentry{app:additional_results}{More Experiments}
\appendixsubentry{app:all_variant_results}{Complete task results}
\appendixsubentry{app:variant_profiles}{Performance profiles across all variants}
\appendixsubentry{app:composition_depth}{Performance by composition depth}
\appendixsubentry{app:engine_transfer}{Transfer across local search engines}
\appendixsubentry{app:routing_adaptation}{VRP target adaptation}
\appendixsubentry{app:scale_generalization}{Generalization to larger VRP instances}
\appendixsubentry{app:heterogeneous_tasks}{Search across heterogeneous tasks}
\appendixsubentry{app:resource_accounting}{Resource use}

\appendixchapterentry{app:prompts}{Generation prompts and interfaces}
\appendixsubentry{app:prompt_templates}{Task descriptions and input templates}
\appendixsubentry{app:prompt_attributes}{Descriptions for active attributes}
\appendixsubentry{app:prompt_instructions}{Generation instructions}

\appendixchapterentry{app:heuristics}{Representative evolved heuristics}
\appendixsubentry{app:heuristic_routing}{VRP example}
\appendixsubentry{app:heuristic_fjsp}{FJSP example}

\appendixchapterentry{app:licenses}{Licenses}
\endgroup
\clearpage

\section{Related work}
\label{sec:related_work}
\subsection{Task-specific automated heuristic design}
\label{app:related_task_specific}

LLM-driven AHD commonly uses evolutionary search to generate and refine executable heuristics from problem-specific feedback \citep{romeraparedes2024funsearch,liu2024eoh,vanstein2025llamea}. Natural-language descriptions and reflection provide additional guidance for program evolution \citep{liu2024eoh,ye2024reevo}. Other search strategies include diversity-aware evolution \citep{dat2025hsevo}, Monte Carlo tree search \citep{zheng2025mctsahd}, and neighborhood search \citep{xie2025lhns}. These approaches differ primarily in how they explore the program space and reuse evaluation feedback.

A complementary direction addresses variation among instances of the same problem. Heuristic-set evolution \citep{liu2025eohs} and instance specialization through feature-based partitioning or performance-based clustering \citep{zhang2026instspechh,hu2026dyca} seek complementary strengths across instance groups. Multi-objective search additionally considers trade-offs between solution quality and computational cost \citep{yao2025meoh}. Applications extend to parameter adaptation and ruin-operator design within vehicle-routing solvers \citep{xie2025mpllm,xie2026ailsahd}, as well as algebraic turbulence-model discovery \citep{zhang2025autoturb}.

\subsection{Cross-task automated heuristic design}
\label{app:related_cross_task}

Cross-task AHD investigates reusable knowledge beyond an individual design task. Existing approaches transfer task-agnostic algorithmic templates through hierarchical representations \citep{liu2026mths}, learn heuristic optimizers through meta-optimization \citep{shi2026moh}, or share program archives before task-specific adaptation \citep{gozeten2026emo}. The reusable knowledge consequently ranges from abstract search principles to executable programs, with downstream feedback guiding their specialization.

\subsection{Knowledge transfer across optimization tasks}
\label{app:related_transfer}

Knowledge transfer across optimization tasks has also been studied through learned solvers and shared search populations. In neural combinatorial optimization, multi-task VRP learning trains a shared solver across problem definitions and studies zero-shot transfer to unseen constraint combinations \citep{liu2024multitaskrouting,zhou2026urs,chen2026space}. Evolutionary multitasking exploits shared representations and cross-task reproduction, with adaptive transfer mechanisms regulating exchange as populations develop \citep{gupta2016multifactorial,bali2020mfea2}. The usefulness of such an exchange motivates studying task relationships. Related work in representation learning characterizes these relationships through Fisher-information embeddings \citep{achille2019task2vec} and directed transfer graphs estimated from predictive performance \citep{zamir2018taskonomy}, distinguishing representational similarity from empirical transfer utility.

\clearpage
\section{More Method Details}
\label{app:implementation}

This section expands the task interactions, population updates, and heuristic-set selection introduced in Section~\ref{sec:method}. We retain the notation $T_C$ for a design task, $\mathcal P_C$ for its population, and $D_{T_j}$ for a source dataset. All search decisions use source evaluations, while target specifications describe the problems that the generated heuristics are intended to solve.

\subsection{Search Procedure}
\label{app:search_allocation}

The explored tasks form a directed acyclic graph. Its roots are the single-attribute tasks, and an interaction between $T_A$ and $T_B$ produces $T_{A\cup B}$. Eligibility requires both $A\subsetneq A\cup B$ and $B\subsetneq A\cup B$, so each parent contributes an attribute absent from the other. Every edge therefore increases the attribute set. Several parent pairs can reach the same task, allowing its population to incorporate heuristics developed through different combinations of prior knowledge.

Algorithm~\ref{alg:meco} summarizes the search under a proposal budget $Q$. Each source receives an initial allowance $q_s$, and each expansion receives a local allowance $q$, with population capacity $N$. A completed interaction retires its parent pair while leaving both parent tasks available for other interactions. Repeatedly reached tasks retain a single merged population. Consequently, task exploration and local evolution share the same budget, and the graph grows according to source feedback.

\par\medskip
\noindent\begin{minipage}{\linewidth}
\refstepcounter{mugealgorithm}\label{alg:meco}
\hrule\vspace{5pt}
\textbf{Algorithm \themugealgorithm\quad \method{} multi-task evolution}
\vspace{4pt}\hrule\vspace{4pt}
\small
\begin{tabbing}
\hspace{1.2em}\=\hspace{1.2em}\=\hspace{1.2em}\=\kill
\textbf{Input}\quad Source datasets $\{D_{T_j}\}_{j=1}^m$, task specifications, proposal budget $Q$\\
\textbf{Input}\quad Allowances $q_s,q$, population capacity $N$, output size $K$\\
Evolve source populations on their respective datasets with allowance $q_s$ each\\
Initialize explored attribute sets $\mathcal V$ with the roots and completed pairs $\mathcal U=\varnothing$\\
\textbf{while} proposal budget remains \textbf{do}\\
\> Form eligible pairs of attribute sets in $\mathcal V$ excluding $\mathcal U$\\
\> \textbf{if} no eligible pair remains \textbf{then break}\\
\> Prefer unexplored unions, then select a pair $(A,B)$ with smallest $G(A,B)$\\
\> Set $C=A\cup B$ and choose the native champion of each parent population\\
\> Initialize a local population $\mathcal P=\varnothing$ and successful-candidate set $\mathcal A_C=\varnothing$\\
\> \textbf{while} local allowance $q$ and total budget remain \textbf{do}\\
\>\> Choose parent programs from the champions initially, then from $\mathcal P$\\
\>\> Generate candidates for $T_C$ with crossover or guided mutation\\
\>\> Evaluate on source datasets and obtain the valid candidate set $\mathcal Q_C$\\
\>\> $\mathcal A_C\gets\mathcal A_C\cup\mathcal Q_C$\\
\>\> $\mathcal P\gets\operatorname{Survive}_C(\mathcal P\cup\mathcal Q_C,N)$\\
\>\> \textbf{if} initialization produces no valid candidate \textbf{then break}\\
\> \textbf{if} $\mathcal A_C\ne\varnothing$ \textbf{then}\\
\>\> Select a retained population $\mathcal S_C$ by $J_C$ and behavioral priority\\
\>\> \textbf{if} $C\in\mathcal V$ \textbf{then} $\mathcal P_C\gets\operatorname{Survive}_C(\mathcal P_C\cup\mathcal S_C,N)$\\
\>\> \textbf{else} set $\mathcal P_C\gets\mathcal S_C$ and add $C$ to $\mathcal V$\\
\>\> Add $(A,B)$ to $\mathcal U$\\
Collect representatives from all task populations\\
Filter repeated behaviors to obtain the eligible pool $\mathcal E$\\
$\mathcal H\gets\{\arg\min_{h\in\mathcal E}m^{-1}\sum_j f_j(h)\}$\\
\textbf{while} $|\mathcal H|<K$ \textbf{do}\\
\> Add the candidate with greatest marginal reduction in $J_{\rm comp}(\mathcal H)$\\
\textbf{Output}\quad Heuristic set $\mathcal H$, frozen before target evaluation
\end{tabbing}
\hrule
\end{minipage}
\par\medskip

The local survival operator preserves source-quality trade-offs and distinct heuristic logic. The retained population then supplies evidence for subsequent task interactions. Final selection draws from all explored tasks. Therefore a heuristic can remain useful even when it was developed for a different task specification.

\subsection{Source Feedback and Transfer}
\label{app:transfer_details}

The normalized score $f_j(h)$ compares a heuristic with a simple reference rule on source $T_j$, as defined in Equation~\ref{eq:local_quality}. Each scale $b_j$ is initially computed on $D_{T_j}$ and kept fixed. This normalization gives each constituent source a comparable contribution to $J_C(h)$. Evaluating a combined-task heuristic produces a complete source vector, while its local quality depends only on the dimensions associated with $C$.

For transfer from $T_A$ to $T_B$, both populations are assessed on exactly the same destination sources. Expanding the definition in the main text gives
\begin{equation}
 s(A\!\rightarrow\!B)=\frac{1}{|\mathcal P_A|\,|B|}\sum_{h\in\mathcal P_A}\sum_{j\,\mid\,c_j\in B}\frac{1}{b_j|D_{T_j}|}\sum_{x\in D_{T_j}}\ell(h,x).
 \label{eq:appendix_transfer_mean}
\end{equation}
The native score $s(B\!\rightarrow\!B)$ is obtained by replacing $\mathcal P_A$ with $\mathcal P_B$ in this expression. Both scores average over all members of their respective populations and all instances in the same destination source datasets. Comparing $s(A\!\rightarrow\!B)$ with $s(B\!\rightarrow\!B)$ gives the directional component $g(A\!\rightarrow\!B)$. Task pairing uses the symmetric transfer gap $G(A,B)$, which averages $g(A\!\rightarrow\!B)$ and $g(B\!\rightarrow\!A)$.

The positive part in Equation~\ref{eq:directional_transfer_gap} measures relative deterioration upon transfer. A direction with no deterioration contributes zero to the symmetric gap. Prioritizing unexplored unions promotes task coverage, while minimizing the gap among equally prioritized pairs favors populations with stronger mutual transfer. Once a population changes, subsequent pair decisions use its updated membership and scores.

\subsection{Heuristic Generation}
\label{app:parent_programs}

An expansion begins with the best native heuristic from each selected task. The target description specifies $T_C$, and the parent descriptions and source scores provide context for transferring their decision rules. Subsequent proposals use representatives of the local population. Grouping programs by their decisions on common construction states prevents behaviorally repeated candidates from dominating parent selection.

The composition and conflict-resolution operators, $\mathrm c_1$ and $\mathrm c_2$, draw distinct parents with a preference for lower $J_C$. For a representative set $\mathcal R_C$, quality-based sampling uses
\begin{equation}
 p_C(h)=\frac{w_C(h)}{\sum_{g\in\mathcal R_C}w_C(g)},
 \qquad
 w_C(h)=\max_{g\in\mathcal R_C}J_C(g)-J_C(h)+\eta,
 \label{eq:appendix_parent_sampling}
\end{equation}
where $\eta>0$ gives every representative a positive weight. We sample the first parent according to $p_C$, remove it from the candidate set, and recompute the probabilities over the remaining candidates to sample the second parent. Composition combines compatible mechanisms, while conflict resolution reconciles competing rules through conditional or hierarchical decisions.

Exploratory crossover $\mathrm c_3$ selects the pair with the greatest difference in constituent-source scores. For distinct candidates $h,g\in\mathcal R_C$, we define their distance and select the parents as
\begin{equation}
 \begin{aligned}
 d_C(h,g)&=\frac{1}{|C|}\sum_{j\,\mid\,c_j\in C}|f_j(h)-f_j(g)|,\\
 (h_1,h_2)&\in\arg\max_{\substack{h,g\in\mathcal R_C\\h\ne g}}d_C(h,g).
 \end{aligned}
 \label{eq:appendix_parent_contrast}
\end{equation}
The selected parents provide contrasting performance profiles, and the generation prompt directs exploration toward an under-covered source dimension. Guided mutation $\mathrm m_1$ uses a strong local parent and its source feedback to revise a decision mechanism while retaining its backbone. These operators act within the task chosen by the outer search and share its source-based evaluation criterion.

\subsection{Population Survival}
\label{app:survival_behavior}

Local survival first compares candidates on their constituent-source vectors. A heuristic $h$ dominates $g$ for $T_C$ when
\begin{equation}
 f_j(h)\leq f_j(g)\quad\forall j\text{ with }c_j\in C,
 \qquad
 f_j(h)<f_j(g)\quad\exists j\text{ with }c_j\in C.
 \label{eq:appendix_dominance}
\end{equation}
Nondominated fronts define the initial ordering. Within a front, crowding distance favors candidates in less densely populated regions of the source-score space \citep{deb2002nsgaii}. This preserves alternative trade-offs among constituent sources during local evolution.

We characterize a heuristic's behavior by its action choices on a fixed set of construction states $\mathcal S$ collected from reference trajectories on source problems. Each state contains the instance information, partial solution, and available actions. In VRP, an action selects the next node to visit. In FJSP, it selects an operation and the machine on which to process it. Every heuristic receives the same states, so its behavior is represented by the action vector $(a(h,s))_{s\in\mathcal S}$, where $a(h,s)$ is the action chosen by $h$ at state $s$. The disagreement rate is
\begin{equation}
 d_{\rm beh}(h,g)=\frac{1}{|\mathcal S|}\sum_{s\in\mathcal S}\mathbf 1\{a(h,s)\ne a(g,s)\}.
 \label{eq:appendix_behavior_distance}
\end{equation}
We scan candidates in the quality order and assign each to the first existing representative within distance $\delta$. A candidate with no such representative starts a new group. Representatives are placed before repeated behaviors, preserving the quality order within each category. Repeated behaviors fill remaining positions when needed. Thus, programs that choose the same actions across the reference states can have distance zero even when their code differs. This comparison follows behavior-aware program comparison \citep{zhang2026rethinking}.

Population retention has two stages: (1) During local evolution, the current members and new candidates are ordered by nondominated fronts and crowding distance, followed by behavioral priority, and at most $N$ survive. (2) After expansion, if $T_C$ is new, these heuristics establish its population. If another parent pair has already produced the same target specification $C$, the existing and newly retained populations are pooled and reduced to at most $N$ members using nondominated sorting, crowding distance, and behavioral priority again.

\subsection{Final Heuristic Set Selection}
\label{app:selection}

We pool candidate heuristics from all explored task populations, including the sources, and sort them by mean normalized source cost. Behavioral grouping then retains the lowest-cost representative of each group to form $\mathcal E$. Multiple candidates from the same task can remain eligible.

We initialize $\mathcal H$ with the candidate having the lowest mean normalized source cost. For each combination $C\in\mathcal C_{\geq2}$, let $q_C(\mathcal H)=\min_{g\in\mathcal H}J_C(g)$ be the current set's best mean score on its constituent sources. Adding a candidate $h\in\mathcal E\setminus\mathcal H$ reduces the set objective by
\begin{equation}
 \Delta(h\mid\mathcal H)=\frac{1}{|\mathcal C_{\geq2}|}\sum_{C\in\mathcal C_{\geq2}}\max\{0,q_C(\mathcal H)-J_C(h)\}.
 \label{eq:appendix_set_gain}
\end{equation}
Only improvements over the current best score contribute to this gain. We add the candidate with the largest $\Delta$, update the combination scores, and repeat until $|\mathcal H|=K$. Ties favor lower mean normalized source cost. The objective covers combinations with $|C|\geq2$, using source scores alone. The selected set is frozen before target evaluation.

\subsection{Constructive decision interface}
\label{app:proposal_accounting}

Within each problem family, programs use a shared constructive decision interface. The current instance and partial solution determine the admissible actions through task-specific feasibility masks. A heuristic ranks or selects among these actions, and construction updates the partial solution accordingly. 

\clearpage
\section{Task definitions and constraint semantics}
\label{app:task_definitions}

We address two families of tasks in this paper. Each contains a base problem and four optional attributes. Active attributes determine the feasibility rules and cost conventions. These models draw on published problems, while their joint instances and augmentation parameters define the task families used in this study.

\subsection{VRP}
\label{app:vrp_constraints}

The base capacitated vehicle routing problem assigns every customer to exactly one route starting at a depot. Vehicles have capacity $Q$, and the objective is total traveled distance. We use the four attributes considered in multi-task VRP solving by \citet{liu2024multitaskrouting}, with the following load and route conventions.

\paragraph{Open routes O.}
A vehicle may end its route at its final customer. The objective excludes the return leg to the depot. This convention models deliveries for which the vehicle need not return to its original dispatch point. It also changes the travel required when checking the remaining route budget.

\paragraph{Time windows TW.}
Service at customer $i$ must start within $[a_i,b_i]$. Early arrivals wait until $a_i$, and departure follows service completion. If $t$ is the departure time from the preceding node and $\tau_i$ is the travel time to $i$, service starts at $\max\{t+\tau_i,a_i\}$. Closed routes also respect the depot return deadline. Time windows represent customer availability and scheduled delivery commitments.

\paragraph{Route-length limits L.}
Each route's total traveled distance is bounded by a prescribed limit. The return leg counts only for closed routes. This bound models a limited travel allowance, such as a distance-based operating range, and remains distinct from waiting and service times under TW.

\paragraph{Mixed deliveries and pickups B.}

B introduces pickups alongside deliveries and permits their visits to interleave. A route whose first customer is a pickup departs empty and serves pickups only. A route beginning with a delivery departs with load $Q$. Let $d_i$ be positive for deliveries and negative for pickups. After visiting customer $v_k$, the load obeys
\begin{equation}
 q_k=q_{k-1}-d_{v_k},\qquad 0\le q_k\le Q.
\end{equation}
The total delivered quantity and total collected quantity on each route are each bounded by $Q$. These conditions remain active when B is combined with O, TW, or L. Customer selection preserves all applicable conditions, and the resulting routes are checked for coverage and feasibility. The attribute represents a vehicle sharing capacity between outbound deliveries and inbound collections. The pickup-first convention separates empty-departure collection routes from routes carrying delivery loads.

\paragraph{Composition and feasibility.}
For a task $P_C$, all attributes in $C$ apply to the same route. For example, O+TW+B combines open-route accounting, service-window checks, and the mixed-load trajectory. At each construction step, a customer is available only if it satisfies every active condition. Capacity and customer coverage remain part of the base problem.

\subsection{Flexible job shop scheduling}
\label{app:fjsp_constraints}

Each job consists of ordered operations. Operation $o$ must be assigned to one eligible machine $m$ and processed without interruption for $p_{om}$ time units. Operations on the same machine cannot overlap. The objective is the makespan, the largest operation completion time.

\paragraph{Release dates R.}
Each job has a release date known before scheduling. Its operations must respect this availability time. The constraint represents the arrival of material or a production order. We specialize the job-arrival setting in \citet{reijnen2026benchmark} to dates available at the start of construction.

\paragraph{Assembly precedence A.}
A sparse acyclic set of cross-job edges augments the within-job operation chains. An operation can begin only after all its predecessors finish. Such edges model components that must be produced before a downstream assembly or dependent processing step. We draw on the coupled-job setting of \citet{lin2022assembly} and generate additional predecessor graphs for our instances.

\paragraph{Sequence-dependent setups S.}
The setup duration depends on the machine and the families of two consecutive operations on that machine. This models changeovers such as tool replacement or cleaning. We follow the sequence-dependent setup semantics studied by \citet{shen2018setup}. The machine is unavailable for processing during its setup, and its first operation uses an initial setup duration.

\paragraph{Transportation delays T.}
Successive operations of a job incur a deterministic delay when assigned to different machines. This represents the time required to move a part between workstations. We adapt the transportation setting of \citet{berterottiere2024transportation} to fixed machine-to-machine lags. The present model imposes no competition for a finite fleet of transport vehicles.

\paragraph{Composition and constructive actions.}

Let $u$ be the preceding operation of $o$ within its job, $m_u$ its assigned machine, $v$ the last operation assigned to $m$, and $\mathcal A(o)$ the active cross-job predecessors. Denote completion times by $C$, the job release date by $r_j$, transportation time by $t$, and setup time by $s$. Construction appends $o$ to machine $m$ at
\begin{equation}
 S_o=\max\left\{r_j,\ C_u+t_{m_u m},\
 \max_{a\in\mathcal A(o)}C_a,\ C_v+s_{mvo}\right\},
 \qquad C_o=S_o+p_{om}.
\end{equation}
Absent predecessors contribute zero, and inactive attributes contribute zero release dates, lags, or setup times. An idle machine may perform setup before its next operation becomes available. The first operation on a machine uses its initial setup time. Transport applies only to within-job edges and is zero between identical machines. An action is eligible once its predecessors have been assigned and its machine is eligible. The time bounds then ensure temporal feasibility.

\clearpage
\section{Benchmark variants and experimental parameters}
\label{app:benchmark_parameters}

\subsection{Complete variant specification}
\label{app:variant_specification}

\tableautorefname~\ref{tab:variant_catalog} enumerates the two task families. Capacity, customer coverage, and a common depot belong to the VRP base problem. Job precedence, machine eligibility, and non-overlapping processing belong to the FJSP base problem. The four attributes in each family are optional additions. Each family therefore contains $1+4+6+4+1=16$ variants. The four single-attribute variants supply training instances. All 16 variants have separate test instances. In terms of task attributes, four task types are in distribution and twelve are held out, comprising the base task and eleven multi-attribute combinations.

\begin{table}[htbp]
 \centering\normalsize
\caption{Benchmark variants and active constraints. ID and OOD denote source and held-out task types.}
 \label{tab:variant_catalog}
 
 \begin{tabular*}{\linewidth}{@{\extracolsep{\fill}}lcccccc@{\hspace{16pt}}lcccccc@{}}
 \toprule
 \multicolumn{7}{c}{VRP} & \multicolumn{7}{c}{FJSP} \\
 \cmidrule(lr){1-7}\cmidrule(lr){8-14}
 Variant & O & B & TW & L & Depth & Split & Variant & R & A & S & T & Depth & Split \\
 \midrule
 Base &  &  &  &  & 0 & OOD & Base &  &  &  &  & 0 & OOD \\
 B &  & $\checkmark$ &  &  & 1 & ID & A &  & $\checkmark$ &  &  & 1 & ID \\
 L &  &  &  & $\checkmark$ & 1 & ID & R & $\checkmark$ &  &  &  & 1 & ID \\
 TW &  &  & $\checkmark$ &  & 1 & ID & S &  &  & $\checkmark$ &  & 1 & ID \\
 O & $\checkmark$ &  &  &  & 1 & ID & T &  &  &  & $\checkmark$ & 1 & ID \\
 BL &  & $\checkmark$ &  & $\checkmark$ & 2 & OOD & AS &  & $\checkmark$ & $\checkmark$ &  & 2 & OOD \\
 BTW &  & $\checkmark$ & $\checkmark$ &  & 2 & OOD & AT &  & $\checkmark$ &  & $\checkmark$ & 2 & OOD \\
 LTW &  &  & $\checkmark$ & $\checkmark$ & 2 & OOD & RA & $\checkmark$ & $\checkmark$ &  &  & 2 & OOD \\
 OB & $\checkmark$ & $\checkmark$ &  &  & 2 & OOD & RS & $\checkmark$ &  & $\checkmark$ &  & 2 & OOD \\
 OL & $\checkmark$ &  &  & $\checkmark$ & 2 & OOD & RT & $\checkmark$ &  &  & $\checkmark$ & 2 & OOD \\
 OTW & $\checkmark$ &  & $\checkmark$ &  & 2 & OOD & ST &  &  & $\checkmark$ & $\checkmark$ & 2 & OOD \\
 BLTW &  & $\checkmark$ & $\checkmark$ & $\checkmark$ & 3 & OOD & AST &  & $\checkmark$ & $\checkmark$ & $\checkmark$ & 3 & OOD \\
 OBL & $\checkmark$ & $\checkmark$ &  & $\checkmark$ & 3 & OOD & RAS & $\checkmark$ & $\checkmark$ & $\checkmark$ &  & 3 & OOD \\
 OBTW & $\checkmark$ & $\checkmark$ & $\checkmark$ &  & 3 & OOD & RAT & $\checkmark$ & $\checkmark$ &  & $\checkmark$ & 3 & OOD \\
 OLTW & $\checkmark$ &  & $\checkmark$ & $\checkmark$ & 3 & OOD & RST & $\checkmark$ &  & $\checkmark$ & $\checkmark$ & 3 & OOD \\
 OBLTW & $\checkmark$ & $\checkmark$ & $\checkmark$ & $\checkmark$ & 4 & OOD & RAST & $\checkmark$ & $\checkmark$ & $\checkmark$ & $\checkmark$ & 4 & OOD \\
 \bottomrule
 \end{tabular*}
\end{table}

\subsection{Problem parameters and data construction}
\label{app:problem_parameters}

All methods use the same source datasets and held-out test instances within each problem family. Training covers the single-attribute tasks, while testing covers both source and unseen task types at a larger instance size. \tableautorefname~\ref{tab:benchmark_sizes} summarizes the data splits.

\begin{table}[htbp]
 \centering\normalsize
 \caption{Dataset sizes and base parameters.}
 \label{tab:benchmark_sizes}
 
 \begin{tabularx}{\linewidth}{@{}p{0.29\linewidth}XX@{}}
 \toprule
 Setting & VRP & FJSP \\
 \midrule
 Training tasks $\times$ instances & $4\times10$ & $4\times10$ \\
 Training size & 50 customers and a depot & 50 operations, 10 jobs, 5 machines \\
 Test tasks $\times$ instances & $16\times1{,}000$ & $16\times1{,}000$ \\
 Test size & 100 customers and a depot & 100 operations, 10 or 20 jobs \\
 Test machines & -- & 5, 10, 11, 12, or 13 \\
 \midrule
 Coordinates / processing data & Uniform coordinates in $[0,1]^2$ & Public processing times and eligibility sets \\
 Demand magnitudes & Uniform in $\{1,\ldots,9\}$ & -- \\
 Vehicle capacity & $Q=50$ & -- \\
 Training templates & -- & 10 Hurink e-data and r-data templates \\
 Validation templates & -- & 10 Hurink s-data and v-data templates \\
 Test templates & -- & 92 Hurink, 7 Barnes, 1 Brandimarte \\
 Realizations per test template & -- & 10 \\
 \bottomrule
 \end{tabularx}
\end{table}

\paragraph{VRP instances.}
Training coordinates and demand magnitudes are sampled independently. Optional attributes add pickups, travel limits, and time windows according to \tableautorefname~\ref{tab:fjsp_augmentation_parameters}. For time windows, the center depends on the customer's distance from the depot, and the sampled interval is clipped to the depot horizon. Instances that fail the single-customer return-feasibility check are regenerated. We use fixed evaluation instances from the public VRP benchmark repository.\footnote{\url{https://github.com/CIAM-Group/URS}} 

\paragraph{FJSP instances.}
We use the public templates provided by \citet{reijnen2026benchmark}, preserving their processing times and eligible-machine sets. Training uses the e-data and r-data versions of Hurink la01 to la05, and validation uses the corresponding s-data and v-data versions. Validation data are excluded from search feedback. Test templates are disjoint from both splits. Each template receives independent realizations of the optional constraints, with each realization shared across all attribute subsets. Thus, variants differ only in which constraints are active. The base variant repeats the unchanged scheduling template across its realizations.

FJSP timing attributes are scaled by the template workload and processing times. It is defined as
\begin{equation}
 H=\max\!\left\{1,\operatorname{round}\!\left(\frac{\sum_o\min_{m\in\mathcal M_o}p_{om}}{M}\right)\right\},
 \qquad
 \bar p=\frac{\sum_o\sum_{m\in\mathcal M_o}p_{om}}{\sum_o|\mathcal M_o|},
\end{equation}
where $M$ is the number of machines, $\mathcal M_o$ contains the eligible machines for operation $o$, and $p_{om}$ is its processing time on machine $m$. \tableautorefname~\ref{tab:fjsp_augmentation_parameters} gives the resulting constraint settings, with $n$ denoting the number of operations. Integer-valued durations are sampled uniformly up to the rounded bound, with a minimum upper bound of one.

\begin{table}[htbp]
 \centering\normalsize
 \caption{Constraint generation parameters.}
 \label{tab:fjsp_augmentation_parameters}
 
 \begin{tabularx}{\linewidth}{@{}llX@{}}
 \toprule
 Family & Attribute & Setting \\
 \midrule
 VRP & O & Routes end at the last customer without returning to the depot \\
 & B & Pickups comprise $20\%$ of customers, chosen uniformly \\
 & L & Route-distance bound of $3$ \\
 & TW & Speed $1$, service duration $s=0.2$, depot horizon $[0,3]$ \\
 & & Window center $\mathcal U(d_i,3-d_i-s)$ and half-width $\mathcal U(s/2,1)$, where $d_i$ is depot distance \\
 \midrule
 FJSP & R & Release dates from $0$ to $0.80H$, with the first job released at zero \\
 & A & From the third job onward, probability $0.15$ of one extra predecessor chosen from earlier jobs \\
 & S & Setup bound $0.60\bar p$, zero within a family, and $F=\min\{5,\max\{3,\operatorname{round}(\sqrt n/2)\}\}$ balanced operation families \\
 & T & Symmetric transportation delays bounded by $1.20\bar p$, with zero delay on the same machine \\
 \bottomrule
 \end{tabularx}
\end{table}

\paragraph{Shared feasibility and task information.}
\label{app:root_realizations}
All methods use the same feasibility masks and receive the same task information, including active-constraint indicators and VRP time-window features. For testing a variant without TW, service durations are zero and every node receives the same nonbinding time window. With $n$ nodes, distance matrix $D$, and speed $v$, its upper bound is $H_{\rm wide}=\max\{1,\max_{i,j}D_{ij}/v\}(5n+21)$. Variants with TW retain their actual windows.

\subsection{Search and baseline parameters}
\label{app:baseline_parameters}

\paragraph{Common protocol.}
Search uses only source-task feedback, and the final heuristic sets are fixed before target evaluation. The common budget counts newly proposed solver programs, including invalid and duplicate proposals. \tableautorefname~\ref{tab:search_parameters} lists the common settings and principal method parameters.

\begin{table}[htbp]
 \centering\normalsize
 \caption{Common search settings and method parameters.}
 \label{tab:search_parameters}
 
 \begin{tabularx}{\linewidth}{@{}p{0.18\linewidth}p{0.18\linewidth}X@{}}
 \toprule
 \multicolumn{2}{@{}l}{Common setting} & Value \\
 \midrule
 \multicolumn{2}{@{}l}{Generation model} & GPT-4o-mini \\
 \multicolumn{2}{@{}l}{New solver proposals per search} & At most 2,000 \\
 \multicolumn{2}{@{}l}{Independent searches / final set size} & 3 / 4 \\
 \midrule
 Method & Population & Principal parameters \\
 \midrule
 EoH & 20 & Initial allowance 40, two crossover parents \\
 ReEvo & 20 & Mutation rate $0.5$, short- and long-term reflection \\
 MCTS-AHD & 10 & Initial population 4, $\alpha=0.5$, $\lambda_0=0.1$ \\
 EoH-S & 10 & First-stage fraction $0.2$, instance-level set selection \\
 MEoH & 20 & Four source objectives, dominance-dissimilarity survival \\
 HSEvo & 10 & Initial allowance 30, mutation rate $0.5$, harmony memory 5 \\
 \midrule
 EMO-STA & 40 & Four islands, archive 16, shared allowance 1,000 and source adaptation $4\times250$ \\
 MoH & \shortstack[l]{Heuristics 10\\Optimizers 10} & At most 10 outer iterations \\
 \midrule
 MECo & 10 retained per task & Source allowance $4\times100$, expansion allowance $16\times100$ \\
 & & Behavioral comparison on 32 states per source with tolerance $0.02$ \\
 \bottomrule
 \end{tabularx}
\end{table}

\paragraph{Final heuristic sets.}
EoH and ReEvo rank their final populations by mean normalized source cost. MCTS-AHD applies the same ranking to feasible programs in its search tree, and HSEvo uses its valid-program archive and final population, including numerical variants. Each selects the best distinct programs up to the prescribed set size.

EoH-S greedily adds the program that most reduces the sum of per-instance best-member losses, starting with the program of lowest mean loss. MEoH pools its final population and elitist archive, ranks candidates in descending order of its native dominance-dissimilarity survival score, and selects four programs. A predetermined random ordering resolves score ties. This four-member extraction adapts the native scoring principle to our common evaluation protocol.

EMO-STA selects each source's final adapted program, and MoH selects each source's best heuristic with its associated optimizer. All selections use training feedback alone.

\subsection{Classical constructive reference rules}
\label{app:classical_rules}

We compare against deterministic construction rules using the same feasibility masks and state transitions as the generated heuristics. Remaining ties are resolved deterministically. \tableautorefname~\ref{tab:classical_constructive} reports their costs on the common test sets.

\paragraph{VRP.}
Nearest-feasible chooses the admissible customer nearest to the current node. Savings-priority begins each route with the farthest admissible customer and then selects the customer maximizing $d(i,0)+d(0,j)-d(i,j)$, where $i$ is the current node, $j$ the candidate, and $0$ the depot. This adapts the savings expression of \citet{clarke1964scheduling} to sequential construction. The expression serves as a selection priority for both closed and open routes. A route ends when no customer is admissible.

\paragraph{FJSP.}
Earliest completion time (ECT) selects the feasible operation and machine assignment with the earliest completion under all active constraints. Most-work-remaining with ECT (MWKR-ECT) first favors the job with the greatest remaining workload,
\begin{equation}
 W_j=\sum_{\substack{o\in j\\o\ \mathrm{unscheduled}}}\min_{m\in\mathcal M_o}p_{om},
\end{equation}
and then uses earliest completion to choose its machine assignment. This extends compound dispatching to the present constraints \citep{lei2022multiaction}. Each deterministic rule is evaluated once, and group means weight variants equally.

\begin{table}[!htbp]
 \centering
 \caption{Classical heuristic costs. Lower is better.}
 \label{tab:classical_constructive}
 \begingroup\normalsize
\begin{tabularx}{\linewidth}{llYYY}
\toprule
Problem & Rule & All (\#16) & ID (\#4) & OOD (\#12) \\
\midrule
VRP & Nearest-feasible & \cellcolor{BestGreen}\MECoNum{00.00}{\mathbf{23.53}} & \MECoNum{00.00}{23.04} & \cellcolor{BestGreen}\MECoNum{00.00}{\mathbf{23.69}} \\
VRP & Savings-priority & \MECoNum{00.00}{24.51} & \cellcolor{BestGreen}\MECoNum{00.00}{\mathbf{22.48}} & \MECoNum{00.00}{25.19} \\
FJSP & ECT & \cellcolor{BestGreen}\MECoNum{0000.0}{\mathbf{1486.8}} & \cellcolor{BestGreen}\MECoNum{0000.0}{\mathbf{1301.4}} & \cellcolor{BestGreen}\MECoNum{0000.0}{\mathbf{1548.6}} \\
FJSP & MWKR(min)-ECT & \MECoNum{0000.0}{1751.7} & \MECoNum{0000.0}{1436.3} & \MECoNum{0000.0}{1856.9} \\
\bottomrule
\end{tabularx}
\par\medskip
\begin{tabularx}{\linewidth}{lYYlYY}
\toprule
VRP & \shortstack{Nearest-\\feasible} & \shortstack{Savings-\\priority} & FJSP & ECT & \shortstack{MWKR\\ECT} \\
\midrule
Base & \MECoNum{00.00}{20.85} & \cellcolor{BestGreen}\MECoNum{00.00}{\mathbf{19.80}} & Base & \cellcolor{BestGreen}\MECoNum{0000.0}{\mathbf{1146.7}} & \MECoNum{0000.0}{1152.3} \\
B & \MECoNum{00.00}{21.98} & \cellcolor{BestGreen}\MECoNum{00.00}{\mathbf{20.33}} & A & \cellcolor{BestGreen}\MECoNum{0000.0}{\mathbf{1324.9}} & \MECoNum{0000.0}{1528.9} \\
L & \MECoNum{00.00}{20.91} & \cellcolor{BestGreen}\MECoNum{00.00}{\mathbf{19.79}} & R & \cellcolor{BestGreen}\MECoNum{0000.0}{\mathbf{1299.8}} & \MECoNum{0000.0}{1500.5} \\
TW & \MECoNum{00.00}{35.61} & \cellcolor{BestGreen}\MECoNum{00.00}{\mathbf{33.90}} & S & \cellcolor{BestGreen}\MECoNum{0000.0}{\mathbf{1291.8}} & \MECoNum{0000.0}{1366.4} \\
O & \cellcolor{BestGreen}\MECoNum{00.00}{\mathbf{13.68}} & \MECoNum{00.00}{15.88} & T & \cellcolor{BestGreen}\MECoNum{0000.0}{\mathbf{1289.2}} & \MECoNum{0000.0}{1349.2} \\
BL & \MECoNum{00.00}{21.43} & \cellcolor{BestGreen}\MECoNum{00.00}{\mathbf{19.44}} & AS & \cellcolor{BestGreen}\MECoNum{0000.0}{\mathbf{1462.2}} & \MECoNum{0000.0}{1725.0} \\
BTW & \MECoNum{00.00}{38.92} & \cellcolor{BestGreen}\MECoNum{00.00}{\mathbf{37.06}} & AT & \cellcolor{BestGreen}\MECoNum{0000.0}{\mathbf{1604.7}} & \MECoNum{0000.0}{1922.7} \\
LTW & \MECoNum{00.00}{35.62} & \cellcolor{BestGreen}\MECoNum{00.00}{\mathbf{33.85}} & RA & \cellcolor{BestGreen}\MECoNum{0000.0}{\mathbf{1464.6}} & \MECoNum{0000.0}{1875.3} \\
OB & \cellcolor{BestGreen}\MECoNum{00.00}{\mathbf{15.62}} & \MECoNum{00.00}{17.82} & RS & \cellcolor{BestGreen}\MECoNum{0000.0}{\mathbf{1429.0}} & \MECoNum{0000.0}{1689.4} \\
OL & \cellcolor{BestGreen}\MECoNum{00.00}{\mathbf{13.68}} & \MECoNum{00.00}{15.80} & RT & \cellcolor{BestGreen}\MECoNum{0000.0}{\mathbf{1469.4}} & \MECoNum{0000.0}{1687.8} \\
OTW & \cellcolor{BestGreen}\MECoNum{00.00}{\mathbf{20.14}} & \MECoNum{00.00}{25.10} & ST & \cellcolor{BestGreen}\MECoNum{0000.0}{\mathbf{1441.4}} & \MECoNum{0000.0}{1540.6} \\
BLTW & \MECoNum{00.00}{38.89} & \cellcolor{BestGreen}\MECoNum{00.00}{\mathbf{36.92}} & AST & \cellcolor{BestGreen}\MECoNum{0000.0}{\mathbf{1744.5}} & \MECoNum{0000.0}{2104.8} \\
OBL & \cellcolor{BestGreen}\MECoNum{00.00}{\mathbf{15.40}} & \MECoNum{00.00}{17.15} & RAS & \cellcolor{BestGreen}\MECoNum{0000.0}{\mathbf{1582.7}} & \MECoNum{0000.0}{2047.9} \\
OBTW & \cellcolor{BestGreen}\MECoNum{00.00}{\mathbf{21.89}} & \MECoNum{00.00}{27.25} & RAT & \cellcolor{BestGreen}\MECoNum{0000.0}{\mathbf{1768.3}} & \MECoNum{0000.0}{2256.9} \\
OLTW & \cellcolor{BestGreen}\MECoNum{00.00}{\mathbf{20.13}} & \MECoNum{00.00}{25.03} & RST & \cellcolor{BestGreen}\MECoNum{0000.0}{\mathbf{1589.6}} & \MECoNum{0000.0}{1859.2} \\
OBLTW & \cellcolor{BestGreen}\MECoNum{00.00}{\mathbf{21.76}} & \MECoNum{00.00}{27.00} & RAST & \cellcolor{BestGreen}\MECoNum{0000.0}{\mathbf{1880.7}} & \MECoNum{0000.0}{2420.8} \\
\bottomrule
\end{tabularx}
\endgroup

\end{table}

\clearpage
\begingroup
\raggedbottom
\section{More Experiments}
\label{app:additional_results}

\subsection{Complete task results}
\label{app:all_variant_results}
\tableautorefname~\ref{tab:routing_all16} and~\ref{tab:fjsp_all16} give the results for every test variant. The four single-constraint variants are ID tasks. The base problem and the eleven multi-constraint combinations are OOD tasks. Best4 selects the lowest whole-dataset mean among four frozen heuristics, as defined in Equation~\ref{eq:test_metrics}. AHD results report means and sample standard deviations over three independent searches, while deterministic rules are evaluated once.

For each method, the main-table rank averages its per-variant midranks over all sixteen variants and all eleven methods. Ties use the displayed precision. The relative gap averages $100(\bar f_{m,v}/\min_k\bar f_{k,v}-1)$ across variants, where $\bar f_{m,v}$ is method $m$'s mean cost on variant $v$. Thus, the reference is the best observed method mean, not a known optimum. Group means and standard deviations are computed by first averaging the relevant variants within each search. In the detailed tables, bold and underlined entries indicate the best and second-best displayed means, and pale green highlights the best entries.

\begin{longtable}{@{}c@{}}
\caption{VRP results on all variants. Lower is better.}\label{tab:routing_all16}\\
\endfirsthead
\multicolumn{1}{l}{Table~\ref{tab:routing_all16} continued} \\
\endhead
\begin{minipage}{\linewidth}\centering
\begin{tabular}{lrrrr}
\toprule
Method & Base & B & L & TW \\
\midrule
Nearest-feasible & \MECoStat{00.00}{0.00}{20.85}{} & \MECoStat{00.00}{0.00}{21.98}{} & \MECoStat{00.00}{0.00}{20.91}{} & \MECoStat{00.00}{0.00}{35.61}{} \\
Savings-priority & \cellcolor{BestGreen}\MECoStat{00.00}{0.00}{\mathbf{19.80}}{} & \cellcolor{BestGreen}\MECoStat{00.00}{0.00}{\mathbf{20.33}}{} & \cellcolor{BestGreen}\MECoStat{00.00}{0.00}{\mathbf{19.79}}{} & \MECoStat{00.00}{0.00}{33.90}{} \\
EoH & \MECoStat{00.00}{0.00}{20.85}{0.00} & \MECoStat{00.00}{0.00}{22.00}{0.04} & \MECoStat{00.00}{0.00}{21.13}{0.24} & \MECoStat{00.00}{0.00}{\underline{31.77}}{0.07} \\
ReEvo & \MECoStat{00.00}{0.00}{20.89}{0.15} & \MECoStat{00.00}{0.00}{22.05}{0.10} & \MECoStat{00.00}{0.00}{20.95}{0.22} & \MECoStat{00.00}{0.00}{32.84}{1.43} \\
MCTS-AHD & \MECoStat{00.00}{0.00}{20.86}{0.00} & \MECoStat{00.00}{0.00}{21.99}{0.03} & \MECoStat{00.00}{0.00}{20.95}{0.06} & \cellcolor{BestGreen}\MECoStat{00.00}{0.00}{\mathbf{31.76}}{0.10} \\
EoH-S & \MECoStat{00.00}{0.00}{20.69}{0.07} & \MECoStat{00.00}{0.00}{22.10}{0.14} & \MECoStat{00.00}{0.00}{20.82}{0.11} & \MECoStat{00.00}{0.00}{32.09}{0.54} \\
MEoH & \MECoStat{00.00}{0.00}{20.86}{0.01} & \MECoStat{00.00}{0.00}{21.98}{0.01} & \MECoStat{00.00}{0.00}{20.94}{0.02} & \MECoStat{00.00}{0.00}{33.53}{0.72} \\
HSEvo & \MECoStat{00.00}{0.00}{20.85}{0.00} & \MECoStat{00.00}{0.00}{21.91}{0.23} & \MECoStat{00.00}{0.00}{20.91}{0.00} & \MECoStat{00.00}{0.00}{32.00}{0.28} \\
EMO-STA & \MECoStat{00.00}{0.00}{20.37}{0.84} & \MECoStat{00.00}{0.00}{21.08}{1.55} & \MECoStat{00.00}{0.00}{20.50}{0.70} & \MECoStat{00.00}{0.00}{35.05}{0.97} \\
MoH & \MECoStat{00.00}{0.00}{\underline{20.25}}{0.76} & \MECoStat{00.00}{0.00}{\underline{21.01}}{1.50} & \MECoStat{00.00}{0.00}{20.39}{0.63} & \MECoStat{00.00}{0.00}{33.19}{0.61} \\
\textbf{MECo} & \MECoStat{00.00}{0.00}{20.37}{0.84} & \MECoStat{00.00}{0.00}{21.08}{1.55} & \MECoStat{00.00}{0.00}{\underline{20.32}}{0.61} & \MECoStat{00.00}{0.00}{31.80}{0.15} \\
\bottomrule
\end{tabular}\end{minipage} \\
\begin{minipage}{\linewidth}\centering
\begin{tabular}{lrrrr}
\toprule
Method & O & BL & BTW & LTW \\
\midrule
Nearest-feasible & \MECoStat{00.00}{0.00}{13.68}{} & \MECoStat{00.00}{0.00}{21.43}{} & \MECoStat{00.00}{0.00}{38.92}{} & \MECoStat{00.00}{0.00}{35.62}{} \\
Savings-priority & \MECoStat{00.00}{0.00}{15.88}{} & \cellcolor{BestGreen}\MECoStat{00.00}{0.00}{\mathbf{19.44}}{} & \MECoStat{00.00}{0.00}{37.06}{} & \MECoStat{00.00}{0.00}{33.85}{} \\
EoH & \MECoStat{00.00}{0.00}{13.68}{0.00} & \MECoStat{00.00}{0.00}{22.27}{0.74} & \MECoStat{00.00}{0.00}{34.80}{0.28} & \MECoStat{00.00}{0.00}{34.00}{2.76} \\
ReEvo & \MECoStat{00.00}{0.00}{13.77}{0.13} & \MECoStat{00.00}{0.00}{21.82}{0.48} & \MECoStat{00.00}{0.00}{35.78}{1.00} & \MECoStat{00.00}{0.00}{33.00}{1.16} \\
MCTS-AHD & \MECoStat{00.00}{0.00}{13.68}{0.00} & \MECoStat{00.00}{0.00}{21.50}{0.12} & \MECoStat{00.00}{0.00}{34.87}{0.43} & \MECoStat{00.00}{0.00}{\underline{31.77}}{0.12} \\
EoH-S & \MECoStat{00.00}{0.00}{13.71}{0.05} & \MECoStat{00.00}{0.00}{21.56}{0.19} & \MECoStat{00.00}{0.00}{35.08}{0.32} & \MECoStat{00.00}{0.00}{32.10}{0.53} \\
MEoH & \MECoStat{00.00}{0.00}{13.68}{0.01} & \MECoStat{00.00}{0.00}{21.46}{0.02} & \MECoStat{00.00}{0.00}{36.20}{0.75} & \MECoStat{00.00}{0.00}{34.27}{0.54} \\
HSEvo & \MECoStat{00.00}{0.00}{13.66}{0.03} & \MECoStat{00.00}{0.00}{21.33}{0.33} & \MECoStat{00.00}{0.00}{\underline{34.78}}{0.90} & \MECoStat{00.00}{0.00}{31.97}{0.16} \\
EMO-STA & \MECoStat{00.00}{0.00}{\underline{13.21}}{0.81} & \MECoStat{00.00}{0.00}{20.72}{1.22} & \MECoStat{00.00}{0.00}{38.29}{1.10} & \MECoStat{00.00}{0.00}{35.08}{0.94} \\
MoH & \cellcolor{BestGreen}\MECoStat{00.00}{0.00}{\mathbf{12.94}}{0.70} & \MECoStat{00.00}{0.00}{\underline{20.55}}{1.10} & \MECoStat{00.00}{0.00}{35.74}{0.55} & \MECoStat{00.00}{0.00}{33.16}{0.61} \\
\textbf{MECo} & \MECoStat{00.00}{0.00}{\underline{13.21}}{0.81} & \MECoStat{00.00}{0.00}{20.58}{1.09} & \cellcolor{BestGreen}\MECoStat{00.00}{0.00}{\mathbf{34.54}}{0.15} & \cellcolor{BestGreen}\MECoStat{00.00}{0.00}{\mathbf{31.74}}{0.10} \\
\bottomrule
\end{tabular}\end{minipage} \\
\begin{minipage}{\linewidth}\centering
\begin{tabular}{lrrrr}
\toprule
Method & OB & OL & OTW & BLTW \\
\midrule
Nearest-feasible & \MECoStat{00.00}{0.00}{15.62}{} & \MECoStat{00.00}{0.00}{13.68}{} & \MECoStat{00.00}{0.00}{20.14}{} & \MECoStat{00.00}{0.00}{38.89}{} \\
Savings-priority & \MECoStat{00.00}{0.00}{17.82}{} & \MECoStat{00.00}{0.00}{15.80}{} & \MECoStat{00.00}{0.00}{25.10}{} & \MECoStat{00.00}{0.00}{36.92}{} \\
EoH & \MECoStat{00.00}{0.00}{15.62}{0.00} & \MECoStat{00.00}{0.00}{14.65}{1.14} & \cellcolor{BestGreen}\MECoStat{00.00}{0.00}{\mathbf{19.65}}{0.37} & \MECoStat{00.00}{0.00}{37.77}{3.01} \\
ReEvo & \MECoStat{00.00}{0.00}{15.78}{0.11} & \MECoStat{00.00}{0.00}{13.83}{0.14} & \MECoStat{00.00}{0.00}{21.24}{1.30} & \MECoStat{00.00}{0.00}{35.92}{0.83} \\
MCTS-AHD & \MECoStat{00.00}{0.00}{15.61}{0.01} & \MECoStat{00.00}{0.00}{13.71}{0.05} & \MECoStat{00.00}{0.00}{20.00}{0.03} & \MECoStat{00.00}{0.00}{34.85}{0.60} \\
EoH-S & \MECoStat{00.00}{0.00}{15.84}{0.41} & \MECoStat{00.00}{0.00}{13.66}{0.03} & \MECoStat{00.00}{0.00}{20.25}{0.42} & \MECoStat{00.00}{0.00}{35.46}{0.99} \\
MEoH & \MECoStat{00.00}{0.00}{15.63}{0.01} & \MECoStat{00.00}{0.00}{13.70}{0.01} & \MECoStat{00.00}{0.00}{\underline{19.83}}{0.13} & \MECoStat{00.00}{0.00}{36.95}{0.59} \\
HSEvo & \MECoStat{00.00}{0.00}{15.57}{0.04} & \MECoStat{00.00}{0.00}{13.67}{0.02} & \MECoStat{00.00}{0.00}{20.02}{0.28} & \MECoStat{00.00}{0.00}{\underline{34.68}}{0.83} \\
EMO-STA & \MECoStat{00.00}{0.00}{\underline{14.64}}{1.69} & \MECoStat{00.00}{0.00}{13.20}{0.83} & \MECoStat{00.00}{0.00}{20.00}{0.25} & \MECoStat{00.00}{0.00}{38.26}{1.08} \\
MoH & \cellcolor{BestGreen}\MECoStat{00.00}{0.00}{\mathbf{14.35}}{1.50} & \MECoStat{00.00}{0.00}{\underline{12.91}}{0.73} & \MECoStat{00.00}{0.00}{20.13}{0.02} & \MECoStat{00.00}{0.00}{35.60}{0.61} \\
\textbf{MECo} & \MECoStat{00.00}{0.00}{\underline{14.64}}{1.69} & \cellcolor{BestGreen}\MECoStat{00.00}{0.00}{\mathbf{12.62}}{0.76} & \MECoStat{00.00}{0.00}{19.94}{0.14} & \cellcolor{BestGreen}\MECoStat{00.00}{0.00}{\mathbf{34.41}}{0.16} \\
\bottomrule
\end{tabular}\end{minipage} \\
\begin{minipage}{\linewidth}\centering
\begin{tabular}{lrrrr}
\toprule
Method & OBL & OBTW & OLTW & OBLTW \\
\midrule
Nearest-feasible & \MECoStat{00.00}{0.00}{15.40}{} & \MECoStat{00.00}{0.00}{21.89}{} & \MECoStat{00.00}{0.00}{20.13}{} & \MECoStat{00.00}{0.00}{21.76}{} \\
Savings-priority & \MECoStat{00.00}{0.00}{17.15}{} & \MECoStat{00.00}{0.00}{27.25}{} & \MECoStat{00.00}{0.00}{25.03}{} & \MECoStat{00.00}{0.00}{27.00}{} \\
EoH & \MECoStat{00.00}{0.00}{16.50}{0.95} & \cellcolor{BestGreen}\MECoStat{00.00}{0.00}{\mathbf{21.42}}{0.54} & \MECoStat{00.00}{0.00}{21.04}{1.86} & \MECoStat{00.00}{0.00}{23.36}{1.90} \\
ReEvo & \MECoStat{00.00}{0.00}{15.78}{0.38} & \MECoStat{00.00}{0.00}{23.10}{1.01} & \MECoStat{00.00}{0.00}{21.27}{1.17} & \MECoStat{00.00}{0.00}{23.13}{0.90} \\
MCTS-AHD & \MECoStat{00.00}{0.00}{15.49}{0.16} & \MECoStat{00.00}{0.00}{21.84}{0.34} & \MECoStat{00.00}{0.00}{19.88}{0.06} & \MECoStat{00.00}{0.00}{21.92}{0.70} \\
EoH-S & \MECoStat{00.00}{0.00}{15.61}{0.28} & \MECoStat{00.00}{0.00}{22.40}{0.26} & \MECoStat{00.00}{0.00}{20.02}{0.57} & \MECoStat{00.00}{0.00}{22.62}{0.50} \\
MEoH & \MECoStat{00.00}{0.00}{15.44}{0.06} & \MECoStat{00.00}{0.00}{\underline{21.45}}{0.10} & \MECoStat{00.00}{0.00}{\underline{19.78}}{0.15} & \MECoStat{00.00}{0.00}{\underline{21.37}}{0.11} \\
HSEvo & \MECoStat{00.00}{0.00}{15.35}{0.08} & \MECoStat{00.00}{0.00}{21.66}{0.91} & \cellcolor{BestGreen}\MECoStat{00.00}{0.00}{\mathbf{19.74}}{0.18} & \cellcolor{BestGreen}\MECoStat{00.00}{0.00}{\mathbf{21.35}}{0.75} \\
EMO-STA & \MECoStat{00.00}{0.00}{14.49}{1.58} & \MECoStat{00.00}{0.00}{21.73}{0.28} & \MECoStat{00.00}{0.00}{19.96}{0.31} & \MECoStat{00.00}{0.00}{21.61}{0.25} \\
MoH & \MECoStat{00.00}{0.00}{\underline{14.18}}{1.39} & \MECoStat{00.00}{0.00}{21.89}{0.00} & \MECoStat{00.00}{0.00}{20.13}{0.00} & \MECoStat{00.00}{0.00}{21.76}{0.00} \\
\textbf{MECo} & \cellcolor{BestGreen}\MECoStat{00.00}{0.00}{\mathbf{13.62}}{0.83} & \MECoStat{00.00}{0.00}{21.59}{0.12} & \MECoStat{00.00}{0.00}{19.83}{0.15} & \MECoStat{00.00}{0.00}{21.45}{0.10} \\
\bottomrule
\end{tabular}\end{minipage} \\
\end{longtable}

\begin{longtable}{@{}c@{}}
\caption{FJSP results on all variants. Lower is better.}\label{tab:fjsp_all16}\\
\endfirsthead
\multicolumn{1}{l}{Table~\ref{tab:fjsp_all16} continued} \\
\endhead
\begin{minipage}{\linewidth}\centering
\begin{tabular}{lrrrr}
\toprule
Method & Base & A & R & S \\
\midrule
ECT & \MECoStat{0000.0}{00.0}{1146.7}{} & \MECoStat{0000.0}{00.0}{1324.9}{} & \MECoStat{0000.0}{00.0}{1299.8}{} & \MECoStat{0000.0}{00.0}{1291.8}{} \\
MWKR-ECT & \MECoStat{0000.0}{00.0}{1152.3}{} & \MECoStat{0000.0}{00.0}{1528.9}{} & \MECoStat{0000.0}{00.0}{1500.5}{} & \MECoStat{0000.0}{00.0}{1366.4}{} \\
EoH & \MECoStat{0000.0}{00.0}{1061.8}{28.3} & \MECoStat{0000.0}{00.0}{1300.4}{34.4} & \MECoStat{0000.0}{00.0}{\underline{1216.5}}{37.0} & \MECoStat{0000.0}{00.0}{1243.0}{57.6} \\
ReEvo & \MECoStat{0000.0}{00.0}{1044.4}{12.1} & \MECoStat{0000.0}{00.0}{1264.1}{17.5} & \MECoStat{0000.0}{00.0}{1245.1}{38.0} & \MECoStat{0000.0}{00.0}{1200.1}{23.7} \\
MCTS-AHD & \MECoStat{0000.0}{00.0}{1049.4}{43.3} & \MECoStat{0000.0}{00.0}{1269.6}{10.2} & \MECoStat{0000.0}{00.0}{1247.5}{87.7} & \MECoStat{0000.0}{00.0}{1214.2}{30.5} \\
EoH-S & \MECoStat{0000.0}{00.0}{1049.7}{40.7} & \MECoStat{0000.0}{00.0}{1258.9}{6.4} & \MECoStat{0000.0}{00.0}{1243.9}{76.2} & \MECoStat{0000.0}{00.0}{1245.8}{37.1} \\
MEoH & \MECoStat{0000.0}{00.0}{1049.2}{32.1} & \MECoStat{0000.0}{00.0}{\underline{1258.2}}{8.5} & \MECoStat{0000.0}{00.0}{1239.5}{27.1} & \MECoStat{0000.0}{00.0}{1231.9}{51.9} \\
HSEvo & \MECoStat{0000.0}{00.0}{\underline{1036.2}}{9.6} & \MECoStat{0000.0}{00.0}{1268.0}{8.9} & \MECoStat{0000.0}{00.0}{1218.3}{11.0} & \MECoStat{0000.0}{00.0}{1215.0}{11.2} \\
EMO-STA & \MECoStat{0000.0}{00.0}{1091.7}{56.0} & \MECoStat{0000.0}{00.0}{1279.8}{37.5} & \MECoStat{0000.0}{00.0}{1259.3}{34.1} & \MECoStat{0000.0}{00.0}{1232.5}{51.5} \\
MoH & \MECoStat{0000.0}{00.0}{1049.4}{2.0} & \MECoStat{0000.0}{00.0}{1264.6}{12.0} & \MECoStat{0000.0}{00.0}{1233.0}{11.5} & \MECoStat{0000.0}{00.0}{\underline{1199.6}}{17.5} \\
\textbf{MECo} & \cellcolor{BestGreen}\MECoStat{0000.0}{00.0}{\mathbf{1035.2}}{5.6} & \cellcolor{BestGreen}\MECoStat{0000.0}{00.0}{\mathbf{1251.8}}{0.7} & \cellcolor{BestGreen}\MECoStat{0000.0}{00.0}{\mathbf{1211.6}}{0.8} & \cellcolor{BestGreen}\MECoStat{0000.0}{00.0}{\mathbf{1181.3}}{2.6} \\
\bottomrule
\end{tabular}\end{minipage} \\
\begin{minipage}{\linewidth}\centering
\begin{tabular}{lrrrr}
\toprule
Method & T & AS & AT & RA \\
\midrule
ECT & \MECoStat{0000.0}{00.0}{1289.2}{} & \MECoStat{0000.0}{00.0}{1462.2}{} & \MECoStat{0000.0}{00.0}{1604.7}{} & \MECoStat{0000.0}{00.0}{1464.6}{} \\
MWKR-ECT & \MECoStat{0000.0}{00.0}{1349.2}{} & \MECoStat{0000.0}{00.0}{1725.0}{} & \MECoStat{0000.0}{00.0}{1922.7}{} & \MECoStat{0000.0}{00.0}{1875.3}{} \\
EoH & \MECoStat{0000.0}{00.0}{1228.3}{23.1} & \MECoStat{0000.0}{00.0}{1464.0}{58.9} & \MECoStat{0000.0}{00.0}{1649.2}{27.8} & \MECoStat{0000.0}{00.0}{1428.7}{30.1} \\
ReEvo & \MECoStat{0000.0}{00.0}{1207.5}{23.2} & \MECoStat{0000.0}{00.0}{1415.2}{26.8} & \MECoStat{0000.0}{00.0}{1589.9}{52.7} & \MECoStat{0000.0}{00.0}{1446.2}{48.2} \\
MCTS-AHD & \MECoStat{0000.0}{00.0}{1227.3}{77.7} & \MECoStat{0000.0}{00.0}{1418.7}{32.9} & \MECoStat{0000.0}{00.0}{1617.5}{50.5} & \MECoStat{0000.0}{00.0}{1416.8}{37.7} \\
EoH-S & \MECoStat{0000.0}{00.0}{1218.0}{64.1} & \MECoStat{0000.0}{00.0}{1439.8}{4.7} & \MECoStat{0000.0}{00.0}{1616.1}{54.1} & \MECoStat{0000.0}{00.0}{\underline{1405.2}}{23.6} \\
MEoH & \MECoStat{0000.0}{00.0}{1219.2}{27.1} & \MECoStat{0000.0}{00.0}{1421.5}{28.1} & \MECoStat{0000.0}{00.0}{1617.0}{40.6} & \MECoStat{0000.0}{00.0}{1428.9}{55.8} \\
HSEvo & \MECoStat{0000.0}{00.0}{\underline{1199.4}}{14.2} & \MECoStat{0000.0}{00.0}{1434.3}{6.1} & \MECoStat{0000.0}{00.0}{1599.4}{43.9} & \MECoStat{0000.0}{00.0}{1431.0}{36.0} \\
EMO-STA & \MECoStat{0000.0}{00.0}{1251.3}{47.2} & \MECoStat{0000.0}{00.0}{1418.8}{34.2} & \MECoStat{0000.0}{00.0}{1580.3}{20.4} & \MECoStat{0000.0}{00.0}{1429.2}{30.4} \\
MoH & \cellcolor{BestGreen}\MECoStat{0000.0}{00.0}{\mathbf{1196.9}}{17.9} & \MECoStat{0000.0}{00.0}{\underline{1393.8}}{10.2} & \MECoStat{0000.0}{00.0}{\underline{1568.9}}{9.4} & \MECoStat{0000.0}{00.0}{1412.5}{11.9} \\
\textbf{MECo} & \MECoStat{0000.0}{00.0}{1200.8}{11.7} & \cellcolor{BestGreen}\MECoStat{0000.0}{00.0}{\mathbf{1388.8}}{1.6} & \cellcolor{BestGreen}\MECoStat{0000.0}{00.0}{\mathbf{1561.9}}{1.0} & \cellcolor{BestGreen}\MECoStat{0000.0}{00.0}{\mathbf{1402.5}}{0.5} \\
\bottomrule
\end{tabular}\end{minipage} \\
\begin{minipage}{\linewidth}\centering
\begin{tabular}{lrrrr}
\toprule
Method & RS & RT & ST & AST \\
\midrule
ECT & \MECoStat{0000.0}{00.0}{1429.0}{} & \MECoStat{0000.0}{00.0}{1469.4}{} & \MECoStat{0000.0}{00.0}{1441.4}{} & \MECoStat{0000.0}{00.0}{1744.5}{} \\
MWKR-ECT & \MECoStat{0000.0}{00.0}{1689.4}{} & \MECoStat{0000.0}{00.0}{1687.8}{} & \MECoStat{0000.0}{00.0}{1540.6}{} & \MECoStat{0000.0}{00.0}{2104.8}{} \\
EoH & \MECoStat{0000.0}{00.0}{1365.3}{43.8} & \MECoStat{0000.0}{00.0}{1423.6}{28.9} & \MECoStat{0000.0}{00.0}{1418.4}{39.5} & \MECoStat{0000.0}{00.0}{1810.1}{44.1} \\
ReEvo & \MECoStat{0000.0}{00.0}{1376.0}{43.9} & \MECoStat{0000.0}{00.0}{1427.4}{27.5} & \MECoStat{0000.0}{00.0}{1358.5}{32.3} & \MECoStat{0000.0}{00.0}{1740.1}{57.0} \\
MCTS-AHD & \MECoStat{0000.0}{00.0}{1385.4}{71.0} & \MECoStat{0000.0}{00.0}{1484.5}{163.7} & \MECoStat{0000.0}{00.0}{1387.3}{67.9} & \MECoStat{0000.0}{00.0}{1755.2}{49.0} \\
EoH-S & \MECoStat{0000.0}{00.0}{1397.5}{65.6} & \MECoStat{0000.0}{00.0}{1464.5}{122.1} & \MECoStat{0000.0}{00.0}{1399.4}{56.7} & \MECoStat{0000.0}{00.0}{1773.5}{34.1} \\
MEoH & \MECoStat{0000.0}{00.0}{1378.2}{43.3} & \MECoStat{0000.0}{00.0}{1441.0}{40.8} & \MECoStat{0000.0}{00.0}{1384.5}{45.3} & \MECoStat{0000.0}{00.0}{1751.5}{51.6} \\
HSEvo & \MECoStat{0000.0}{00.0}{1358.0}{12.9} & \MECoStat{0000.0}{00.0}{\underline{1415.6}}{31.6} & \MECoStat{0000.0}{00.0}{1365.9}{10.9} & \MECoStat{0000.0}{00.0}{1755.8}{29.4} \\
EMO-STA & \MECoStat{0000.0}{00.0}{1372.7}{48.1} & \MECoStat{0000.0}{00.0}{1441.9}{18.7} & \MECoStat{0000.0}{00.0}{1388.2}{45.6} & \MECoStat{0000.0}{00.0}{1722.1}{23.9} \\
MoH & \MECoStat{0000.0}{00.0}{\underline{1345.4}}{14.6} & \MECoStat{0000.0}{00.0}{1417.7}{12.7} & \MECoStat{0000.0}{00.0}{\underline{1349.0}}{23.7} & \MECoStat{0000.0}{00.0}{\underline{1699.2}}{12.4} \\
\textbf{MECo} & \cellcolor{BestGreen}\MECoStat{0000.0}{00.0}{\mathbf{1326.9}}{1.0} & \cellcolor{BestGreen}\MECoStat{0000.0}{00.0}{\mathbf{1393.9}}{1.6} & \cellcolor{BestGreen}\MECoStat{0000.0}{00.0}{\mathbf{1334.6}}{1.2} & \cellcolor{BestGreen}\MECoStat{0000.0}{00.0}{\mathbf{1692.2}}{5.7} \\
\bottomrule
\end{tabular}\end{minipage} \\
\begin{minipage}{\linewidth}\centering
\begin{tabular}{lrrrr}
\toprule
Method & RAS & RAT & RST & RAST \\
\midrule
ECT & \MECoStat{0000.0}{00.0}{1582.7}{} & \MECoStat{0000.0}{00.0}{1768.3}{} & \MECoStat{0000.0}{00.0}{1589.6}{} & \MECoStat{0000.0}{00.0}{1880.7}{} \\
MWKR-ECT & \MECoStat{0000.0}{00.0}{2047.9}{} & \MECoStat{0000.0}{00.0}{2256.9}{} & \MECoStat{0000.0}{00.0}{1859.2}{} & \MECoStat{0000.0}{00.0}{2420.8}{} \\
EoH & \MECoStat{0000.0}{00.0}{1561.4}{40.8} & \MECoStat{0000.0}{00.0}{1799.7}{22.7} & \MECoStat{0000.0}{00.0}{1570.6}{27.0} & \MECoStat{0000.0}{00.0}{1927.8}{23.2} \\
ReEvo & \MECoStat{0000.0}{00.0}{1574.4}{53.3} & \MECoStat{0000.0}{00.0}{1782.3}{58.8} & \MECoStat{0000.0}{00.0}{1554.0}{43.2} & \MECoStat{0000.0}{00.0}{1910.3}{69.1} \\
MCTS-AHD & \MECoStat{0000.0}{00.0}{1541.1}{35.8} & \MECoStat{0000.0}{00.0}{1786.0}{69.0} & \MECoStat{0000.0}{00.0}{1612.6}{147.8} & \MECoStat{0000.0}{00.0}{1896.3}{53.6} \\
EoH-S & \MECoStat{0000.0}{00.0}{1556.9}{25.9} & \MECoStat{0000.0}{00.0}{1786.0}{71.5} & \MECoStat{0000.0}{00.0}{1600.4}{108.6} & \MECoStat{0000.0}{00.0}{1908.2}{49.3} \\
MEoH & \MECoStat{0000.0}{00.0}{1555.5}{58.9} & \MECoStat{0000.0}{00.0}{1801.2}{76.2} & \MECoStat{0000.0}{00.0}{1562.2}{54.8} & \MECoStat{0000.0}{00.0}{1909.0}{80.6} \\
HSEvo & \MECoStat{0000.0}{00.0}{1561.8}{32.1} & \MECoStat{0000.0}{00.0}{1781.4}{83.6} & \MECoStat{0000.0}{00.0}{1545.9}{22.6} & \MECoStat{0000.0}{00.0}{1907.2}{68.6} \\
EMO-STA & \MECoStat{0000.0}{00.0}{1543.2}{36.0} & \MECoStat{0000.0}{00.0}{1743.6}{21.2} & \MECoStat{0000.0}{00.0}{1549.6}{41.5} & \MECoStat{0000.0}{00.0}{1856.6}{24.6} \\
MoH & \MECoStat{0000.0}{00.0}{\underline{1523.2}}{11.3} & \MECoStat{0000.0}{00.0}{\underline{1729.2}}{14.6} & \MECoStat{0000.0}{00.0}{\underline{1521.4}}{19.3} & \MECoStat{0000.0}{00.0}{\underline{1838.3}}{15.1} \\
\textbf{MECo} & \cellcolor{BestGreen}\MECoStat{0000.0}{00.0}{\mathbf{1511.9}}{2.5} & \cellcolor{BestGreen}\MECoStat{0000.0}{00.0}{\mathbf{1717.9}}{1.4} & \cellcolor{BestGreen}\MECoStat{0000.0}{00.0}{\mathbf{1500.4}}{2.0} & \cellcolor{BestGreen}\MECoStat{0000.0}{00.0}{\mathbf{1827.1}}{3.0} \\
\bottomrule
\end{tabular}\end{minipage} \\
\end{longtable}

\subsection{Performance profiles across all variants}
\label{app:variant_profiles}

Figure~\ref{fig:variant_radars} presents the per-variant relative gaps defined above. The radius uses $\log(1+g)$, where $g$ is the percentage gap, and the tick labels retain the original percentages. This scale makes small differences visible. A smaller radius indicates a lower cost relative to the best observed method. The shaded regions identify methods rather than uncertainty, and their areas should not be interpreted as an aggregate score.

\begin{figure}[!htbp]
 \centering
 \includegraphics[width=\linewidth]{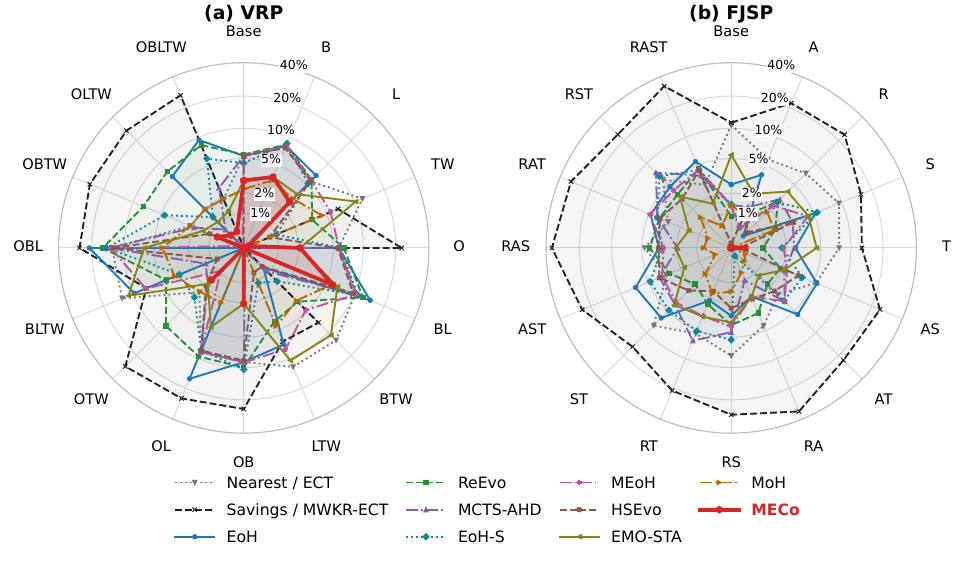}
 \caption{Relative gaps across all variants. Lower is better. MECo is red.}
 \label{fig:variant_radars}
\end{figure}

\subsection{Performance by composition depth}
\label{app:composition_depth}

Composition depth is the number of active optional constraints. Figure~\ref{fig:depth} and Table~\ref{tab:composition_depth} summarize the multi-constraint variants at each depth. Costs are first averaged within each search and depth, then summarized across the three searches. The figure shows selected methods for readability, while the table includes the complete comparison. MECo leads at depths two and three in VRP and at all reported depths in FJSP, while HSEvo has the lowest mean on the four-constraint VRP variant.

\begin{figure}[!htbp]
 \centering
 \includegraphics[width=\linewidth]{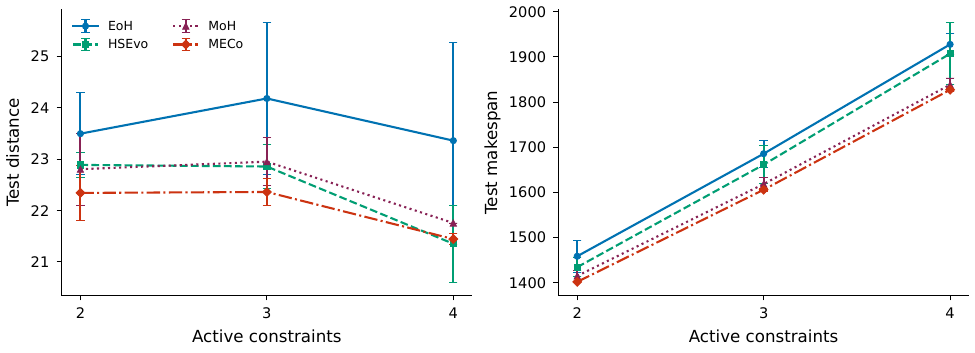}
 \caption{Costs by composition depth, with sample standard deviations.}
 \label{fig:depth}
\end{figure}
\begin{papertable}
 \centering
 \caption{Costs by composition depth. Lower is better.}
 \label{tab:composition_depth}
 \begin{tabularx}{\linewidth}{lYYYYYY}
\toprule
& \multicolumn{3}{c}{VRP} & \multicolumn{3}{c}{FJSP} \\
\cmidrule(lr){2-4}\cmidrule(lr){5-7}
Method & Depth 2 & Depth 3 & Depth 4 & Depth 2 & Depth 3 & Depth 4 \\
\midrule
\shortstack[l]{Nearest-feasible (VRP)\\ECT (FJSP)} & \MECoDepthStat{00.00}{24.23}{} & \MECoDepthStat{00.00}{24.08}{} & \MECoDepthStat{00.00}{21.76}{} & \MECoDepthStat{0000.0}{1478.5}{} & \MECoDepthStat{0000.0}{1671.3}{} & \MECoDepthStat{0000.0}{1880.7}{} \\
\shortstack[l]{Savings-priority (VRP)\\MWKR-ECT (FJSP)} & \MECoDepthStat{00.00}{24.85}{} & \MECoDepthStat{00.00}{26.59}{} & \MECoDepthStat{00.00}{27.00}{} & \MECoDepthStat{0000.0}{1740.1}{} & \MECoDepthStat{0000.0}{2067.2}{} & \MECoDepthStat{0000.0}{2420.8}{} \\
EoH & \MECoDepthStat{00.00}{23.50}{0.80} & \MECoDepthStat{00.00}{24.18}{1.47} & \MECoDepthStat{00.00}{23.36}{1.90} & \MECoDepthStat{0000.0}{1458.2}{35.7} & \MECoDepthStat{0000.0}{1685.5}{30.2} & \MECoDepthStat{0000.0}{1927.8}{23.2} \\
ReEvo & \MECoDepthStat{00.00}{23.57}{0.55} & \MECoDepthStat{00.00}{24.02}{0.74} & \MECoDepthStat{00.00}{23.13}{0.90} & \MECoDepthStat{0000.0}{1435.5}{29.4} & \MECoDepthStat{0000.0}{1662.7}{47.3} & \MECoDepthStat{0000.0}{1910.3}{69.1} \\
MCTS-AHD & \MECoDepthStat{00.00}{22.91}{0.12} & \MECoDepthStat{00.00}{23.01}{0.29} & \MECoDepthStat{00.00}{21.92}{0.70} & \MECoDepthStat{0000.0}{1451.7}{53.6} & \MECoDepthStat{0000.0}{1673.7}{51.3} & \MECoDepthStat{0000.0}{1896.3}{53.6} \\
EoH-S & \MECoDepthStat{00.00}{23.08}{0.15} & \MECoDepthStat{00.00}{23.37}{0.17} & \MECoDepthStat{00.00}{22.62}{0.50} & \MECoDepthStat{0000.0}{1453.8}{48.4} & \MECoDepthStat{0000.0}{1679.2}{53.2} & \MECoDepthStat{0000.0}{1908.2}{49.3} \\
MEoH & \MECoDepthStat{00.00}{23.52}{0.04} & \MECoDepthStat{00.00}{23.40}{0.22} & \MECoDepthStat{00.00}{21.37}{0.11} & \MECoDepthStat{0000.0}{1445.2}{34.4} & \MECoDepthStat{0000.0}{1667.6}{54.4} & \MECoDepthStat{0000.0}{1909.0}{80.6} \\
HSEvo & \MECoDepthStat{00.00}{22.89}{0.24} & \MECoDepthStat{00.00}{22.86}{0.43} & \cellcolor{BestGreen}\MECoDepthStat{00.00}{\mathbf{21.35}}{0.75} & \MECoDepthStat{0000.0}{1434.0}{20.7} & \MECoDepthStat{0000.0}{1661.2}{41.8} & \MECoDepthStat{0000.0}{1907.2}{68.6} \\
EMO-STA & \MECoDepthStat{00.00}{23.66}{0.54} & \MECoDepthStat{00.00}{23.61}{0.41} & \MECoDepthStat{00.00}{21.61}{0.25} & \MECoDepthStat{0000.0}{1438.5}{32.4} & \MECoDepthStat{0000.0}{1639.6}{30.2} & \MECoDepthStat{0000.0}{1856.6}{24.6} \\
MoH & \MECoDepthStat{00.00}{22.81}{0.70} & \MECoDepthStat{00.00}{22.95}{0.47} & \MECoDepthStat{00.00}{21.76}{0.00} & \MECoDepthStat{0000.0}{1414.6}{12.5} & \MECoDepthStat{0000.0}{1618.2}{14.0} & \MECoDepthStat{0000.0}{1838.3}{15.1} \\
\textbf{MECo} & \cellcolor{BestGreen}\MECoDepthStat{00.00}{\mathbf{22.34}}{0.54} & \cellcolor{BestGreen}\MECoDepthStat{00.00}{\mathbf{22.36}}{0.26} & \MECoDepthStat{00.00}{21.45}{0.10} & \cellcolor{BestGreen}\MECoDepthStat{0000.0}{\mathbf{1401.4}}{0.7} & \cellcolor{BestGreen}\MECoDepthStat{0000.0}{\mathbf{1605.6}}{2.9} & \cellcolor{BestGreen}\MECoDepthStat{0000.0}{\mathbf{1827.1}}{3.0} \\
\bottomrule
\end{tabularx}

\end{papertable}

\subsection{Transferring the framework across local search engines}
\label{app:engine_transfer}

The framework study retains each engine's native program generation and survival rules, while MECo supplies the task populations, task pairing, and final heuristic-set selection. No additional execution-error repair is introduced. The main comparison in Table~\ref{tab:engine_transfer} reports three searches for each engine and family. If transferred parents cannot support native initialization, the hybrids use the engine's initialization procedure. For HSEvo, distinct programs with equal fitness may serve as parents when its usual selection rule cannot proceed.

For the additional FJSP reliability study, Table~\ref{tab:fjsp_engine_runs} lists ten independent searches for EoH, ReEvo, and their MECo counterparts. Each value averages a search's costs over the ID or OOD variants. The ten searches include the initial three and seven additional searches. We separately compare EoH with EoH-MECo and ReEvo with ReEvo-MECo on each split, giving four tests.

Each test is a two-sided exact independent permutation test. Its statistic is the absolute difference between the two groups' mean costs, and its null distribution enumerates all $\binom{20}{10}$ assignments of the twenty searches to groups of ten. Holm adjustment is then applied jointly to the four resulting $p$-values. Searches, rather than individual instances, are the independent observations. The adjusted values are reported on the MECo rows. All four are below $0.05$, supporting the lower ID and OOD costs observed in Figure~\ref{fig:fjsp_engine_significance}.

\begin{papertable}
 \centering
 \caption{FJSP costs over ten searches and Holm-adjusted $p$-values.}
 \label{tab:fjsp_engine_runs}
 \begin{tabular*}{\linewidth}{@{\extracolsep{\fill}}lrrrrrr@{}}
\toprule
Method & Run 1 & Run 2 & Run 3 & Run 4 & Run 5 & $p$ \\
\midrule
\multicolumn{7}{@{}l}{\textbf{FJSP ID (\#4)}} \\
EoH & 1203.90 & 1273.08 & 1264.14 & 1226.42 & 1216.71 & -- \\
EoH-MECo & 1226.10 & 1206.31 & 1200.84 & 1212.34 & 1214.94 & 0.01146 \\
ReEvo & 1229.89 & 1216.86 & 1240.74 & 1213.55 & 1232.26 & -- \\
ReEvo-MECo & 1189.72 & 1225.70 & 1206.62 & 1208.58 & 1230.63 & 0.01146 \\
\midrule
\multicolumn{7}{@{}l}{\textbf{FJSP OOD (\#12)}} \\
EoH & 1503.36 & 1563.39 & 1553.41 & 1499.68 & 1512.43 & -- \\
EoH-MECo & 1508.95 & 1508.01 & 1500.05 & 1509.39 & 1493.24 & 0.00351 \\
ReEvo & 1534.43 & 1477.15 & 1543.06 & 1477.61 & 1521.41 & -- \\
ReEvo-MECo & 1463.86 & 1522.43 & 1469.90 & 1469.16 & 1523.56 & 0.01146 \\
\bottomrule
\end{tabular*}
\par\medskip
\begin{tabular*}{\linewidth}{@{\extracolsep{\fill}}lrrrrrr@{}}
\toprule
Method & Run 6 & Run 7 & Run 8 & Run 9 & Run 10 & $p$ \\
\midrule
\multicolumn{7}{@{}l}{\textbf{FJSP ID (\#4)}} \\
EoH & 1344.73 & 1242.49 & 1231.08 & 1210.22 & 1284.11 & -- \\
EoH-MECo & 1212.56 & 1215.72 & 1216.31 & 1214.25 & 1213.56 & 0.01146 \\
ReEvo & 1284.63 & 1218.15 & 1208.12 & 1248.48 & 1323.22 & -- \\
ReEvo-MECo & 1158.15 & 1204.30 & 1197.26 & 1215.17 & 1212.56 & 0.01146 \\
\midrule
\multicolumn{7}{@{}l}{\textbf{FJSP OOD (\#12)}} \\
EoH & 1722.40 & 1541.62 & 1533.27 & 1503.76 & 1627.10 & -- \\
EoH-MECo & 1475.81 & 1497.05 & 1497.31 & 1493.51 & 1485.92 & 0.00351 \\
ReEvo & 1576.14 & 1492.95 & 1490.40 & 1552.26 & 1692.36 & -- \\
ReEvo-MECo & 1441.29 & 1467.63 & 1460.41 & 1477.38 & 1474.04 & 0.01146 \\
\bottomrule
\end{tabular*}
\par\smallskip\noindent Both panels repeat the same $p$-values, computed from all ten searches.

\end{papertable}

\subsection{VRP target adaptation}
\label{app:routing_adaptation}

We further allow the frozen heuristic sets to adapt using training data from two target tasks, OB and BLTW. Each task provides ten size-50 training instances, and evaluation uses the unchanged size-100 test set. Local baselines search independently on each target, whereas EMO-STA, MoH, and MECo share their additional proposal allowance across the two targets. Table~\ref{tab:routing_adaptation} reports the final costs and the total allowance for both targets.

All final programs are selected using target training feedback, and Best4 is computed only after selection. The entries report means and sample standard deviations over three searches. Average rank is the mean of the two target ranks among all listed methods, with ties assigned midranks at the displayed precision. EMO-STA uses its native scoring mechanism. Although MECo does not achieve the lowest cost on either target, its costs remain close to the best observed values and it achieves the best average rank. Moreover, MECo uses half as many additional proposals as the task-specific baselines, demonstrating more efficient use of the adaptation proposal budget.

\begin{papertable}
 \centering
 \caption{VRP target adaptation. Costs are mean $\pm$ sample SD over three searches.}
 \label{tab:routing_adaptation}
 \begin{tabularx}{\linewidth}{lYYYY}
\toprule
Method & Total proposals & OB $\downarrow$ & BLTW $\downarrow$ & Avg. rank $\downarrow$ \\
\midrule
EoH & $500+500$ & \MECoStack{15.7304}{0.1173} & \MECoStack{34.1699}{0.4925} & $5.00$ \\
EoH-S & $500+500$ & \MECoStack{15.5708}{0.1741} & \MECoStack{34.1896}{0.4038} & $4.50$ \\
ReEvo & $500+500$ & \MECoStack{15.7903}{0.0376} & \cellcolor{BestGreen}\MECoStack{\mathbf{33.8441}}{0.0562} & $4.00$ \\
MEoH & $500+500$ & \MECoStack{15.7702}{0.0267} & \MECoStack{34.1351}{0.5875} & $5.00$ \\
MCTS-AHD & $500+500$ & \MECoStack{15.8940}{0.4708} & \MECoStack{34.5688}{0.0305} & $8.00$ \\
HSEvo & $500+500$ & \MECoStack{15.9109}{0.1988} & \MECoStack{34.1125}{0.4006} & $6.00$ \\
\midrule
EMO-STA & $500$ & \MECoStack{15.6168}{0.0000} & \MECoStack{38.8891}{0.0000} & $6.50$ \\
MoH & $500$ & \cellcolor{BestGreen}\MECoStack{\mathbf{14.3453}}{1.5030} & \MECoStack{34.2801}{0.3966} & $4.00$ \\
\midrule
\textbf{MECo} & $500$ & \MECoStack{14.6400}{1.6918} & \MECoStack{33.8955}{0.1653} & \cellcolor{BestGreen}$\mathbf{2.00}$ \\
\bottomrule
\end{tabularx}

\end{papertable}

\subsection{Generalization to larger VRP instances}
\label{app:scale_generalization}

We evaluate the frozen MECo sets on 1,000-customer VRP instances and compare them with the public 100-customer MTNCO checkpoint \citep{liu2024multitaskrouting}. Both methods use the same 100 test instances per variant. MTNCO uses 1,000 starting points, either without augmentation or with eight augmentations. MECo retains whole-dataset Best4 evaluation and reports variation across its three searches. This is a scale-generalization comparison, not an equal-training-budget comparison. ID and OOD follow MECo's source-task split and do not imply that all OOD variants were unseen during MTNCO training. As shown in Table~\ref{tab:vrp_scale1000}, MECo has lower aggregate costs in both groups.

\begin{papertable}
 \centering
 \caption{Frozen evaluation on 1,000-customer VRP instances. Lower is better.}
 \label{tab:vrp_scale1000}
 \begin{tabularx}{\linewidth}{lYYY}
\toprule
Method & All (\#16) & ID (\#4) & OOD (\#12) \\
\midrule
MTN & $148.67$ & $115.16$ & $159.85$ \\
MTN 8aug & $137.05$ & $109.33$ & $146.29$ \\
\rowcolor{BestGreen}\textbf{MECo} & $\mathbf{106.60}\pm 2.54$ & $\mathbf{88.75}\pm 2.40$ & $\mathbf{112.55}\pm 2.62$ \\
\bottomrule
\end{tabularx}

\end{papertable}

\subsection{Search across heterogeneous tasks}
\label{app:heterogeneous_tasks}

We also study joint heuristic design across TSP, base VRP, and FJSP. Each method shares an allowance of 1,000 new programs across the three tasks and selects one executable expert per task using training feedback. Each task uses 64 training instances of size 50 and 1,000 test instances at each of sizes 50 and 100, measured in nodes, customers, or operations. The same expert is evaluated at both test sizes. Table~\ref{tab:heterogeneous_tasks} reports means and sample standard deviations over three searches, without averaging incompatible objective units.

MECo achieves lower TSP and VRP costs than EMO-STA and MoH at both test sizes. It also improves over EMO-STA on FJSP, where MoH performs best. These results extend the evaluation of MECo from constraint variants within a family to joint design across different constructive problems. The experiment concerns the participating tasks, rather than zero-shot transfer to an unseen problem type.

\begin{papertable}
 \centering
 \caption{Joint design across three tasks. Mean $\pm$ sample SD.}
 \label{tab:heterogeneous_tasks}
 \begin{tabularx}{\linewidth}{lrYYY}
\toprule
Task & Size & EMO-STA & MoH & MECo \\
\midrule
\multicolumn{5}{l}{\textbf{3 tasks}} \\
TSP & 50 & \MECoStack{6.99}{0.00} & \MECoStack{6.46}{0.19} & \cellcolor{BestGreen}\MECoStack{\mathbf{6.34}}{0.07} \\
 & 100 & \MECoStack{9.69}{0.00} & \MECoStack{9.04}{0.32} & \cellcolor{BestGreen}\MECoStack{\mathbf{8.83}}{0.11} \\
VRP & 50 & \MECoStack{13.53}{0.00} & \MECoStack{13.41}{0.63} & \cellcolor{BestGreen}\MECoStack{\mathbf{12.97}}{0.51} \\
 & 100 & \MECoStack{23.35}{0.00} & \MECoStack{23.03}{1.16} & \cellcolor{BestGreen}\MECoStack{\mathbf{22.31}}{1.02} \\
FJSP & 50 & \MECoStack{132.87}{7.03} & \cellcolor{BestGreen}\MECoStack{\mathbf{115.02}}{0.30} & \MECoStack{123.00}{9.15} \\
 & 100 & \MECoStack{236.38}{12.29} & \cellcolor{BestGreen}\MECoStack{\mathbf{210.38}}{0.92} & \MECoStack{222.57}{11.72} \\
\bottomrule
\end{tabularx}

\end{papertable}

\subsection{Resource use}
\label{app:resource_accounting}

Table~\ref{tab:resource_audit} reports training time, test time, and token use for the main experiments. Training includes program generation, source evaluation, and final selection. Testing covers the frozen evaluation.

\begin{papertable}
 \centering
 \caption{Resource use per search, with sample standard deviations.}
 \label{tab:resource_audit}
 \begin{tabularx}{\linewidth}{lYYYYY}
\toprule
Method & Train (h) & Test (h) & Input (M) & Output (M) & Tokens (M) \\
\midrule
\multicolumn{6}{l}{\textbf{VRP}} \\
EoH & \MECoStack{0.77}{0.26} & \MECoStack{1.26}{0.13} & \MECoStack{2.16}{0.09} & \MECoStack{0.84}{0.03} & \MECoStack{3.00}{0.11} \\
ReEvo & \MECoStack{1.04}{0.08} & \MECoStack{0.87}{0.17} & \MECoStack{4.97}{0.17} & \MECoStack{1.26}{0.07} & \MECoStack{6.23}{0.23} \\
MCTS-AHD & \MECoStack{5.64}{0.07} & \MECoStack{0.94}{0.16} & \MECoStack{8.17}{0.55} & \MECoStack{1.18}{0.02} & \MECoStack{9.35}{0.56} \\
EoH-S & \MECoStack{0.99}{0.32} & \MECoStack{1.25}{0.18} & \MECoStack{2.54}{0.06} & \MECoStack{1.03}{0.04} & \MECoStack{3.57}{0.11} \\
MEoH & \MECoStack{0.92}{0.29} & \MECoStack{1.25}{0.09} & \MECoStack{2.21}{0.18} & \MECoStack{0.88}{0.05} & \MECoStack{3.09}{0.22} \\
HSEvo & \MECoStack{5.17}{0.67} & \MECoStack{1.06}{0.08} & \MECoStack{5.72}{0.36} & \MECoStack{1.39}{0.13} & \MECoStack{7.11}{0.49} \\
EMO-STA & \MECoStack{3.90}{0.29} & \MECoStack{1.48}{0.10} & \MECoStack{12.05}{1.22} & \MECoStack{0.92}{0.09} & \MECoStack{12.97}{1.32} \\
MoH & \MECoStack{1.39}{0.48} & \MECoStack{1.55}{0.43} & \MECoStack{1.47}{0.89} & \MECoStack{1.11}{0.66} & \MECoStack{2.58}{1.55} \\
MECo & \MECoStack{2.26}{0.12} & \MECoStack{0.98}{0.20} & \MECoStack{3.41}{0.06} & \MECoStack{1.27}{0.03} & \MECoStack{4.68}{0.09} \\
\midrule
\multicolumn{6}{l}{\textbf{FJSP}} \\
EoH & \MECoStack{0.97}{0.15} & \MECoStack{0.09}{0.01} & \MECoStack{2.57}{0.08} & \MECoStack{1.07}{0.04} & \MECoStack{3.64}{0.12} \\
ReEvo & \MECoStack{1.01}{0.21} & \MECoStack{0.06}{0.01} & \MECoStack{7.13}{0.91} & \MECoStack{1.86}{0.31} & \MECoStack{8.99}{1.22} \\
MCTS-AHD & \MECoStack{7.32}{2.57} & \MECoStack{0.04}{0.00} & \MECoStack{11.66}{1.92} & \MECoStack{1.63}{0.08} & \MECoStack{13.28}{1.99} \\
EoH-S & \MECoStack{0.66}{0.02} & \MECoStack{0.06}{0.04} & \MECoStack{2.82}{0.10} & \MECoStack{1.16}{0.05} & \MECoStack{3.98}{0.15} \\
MEoH & \MECoStack{0.82}{0.17} & \MECoStack{0.07}{0.02} & \MECoStack{2.69}{0.30} & \MECoStack{1.09}{0.14} & \MECoStack{3.77}{0.43} \\
HSEvo & \MECoStack{2.46}{0.63} & \MECoStack{0.06}{0.02} & \MECoStack{7.14}{0.10} & \MECoStack{1.64}{0.07} & \MECoStack{8.78}{0.18} \\
EMO-STA & \MECoStack{6.25}{0.35} & \MECoStack{0.09}{0.01} & \MECoStack{23.52}{1.94} & \MECoStack{1.69}{0.14} & \MECoStack{25.21}{2.08} \\
MoH & \MECoStack{2.82}{0.18} & \MECoStack{0.26}{0.21} & \MECoStack{3.51}{0.09} & \MECoStack{2.01}{0.17} & \MECoStack{5.51}{0.20} \\
MECo & \MECoStack{1.79}{0.04} & \MECoStack{0.04}{0.01} & \MECoStack{4.35}{0.02} & \MECoStack{1.33}{0.01} & \MECoStack{5.68}{0.03} \\
\bottomrule
\end{tabularx}

\end{papertable}

\clearpage
\endgroup
\section{Generation prompts and interfaces}
\label{app:prompts}

A generation prompt combines a task description, a function interface, the descriptions of active constraints, and an operator instruction. Parent programs and their source scores provide the search context. We present these components in that order.

\begingroup
\raggedbottom

\subsection{Task descriptions and input templates}
\label{app:prompt_templates}

The VRP contract below comes from the active-constraint-aware search interface. Its accompanying template specifies the five function arguments.

\begin{promptbox}{VRP task description}
Choose one next node for the active route. `instance['active_constraints']` lists the lowercase constraint primitives active in the current variant; compose behavior from primitive membership instead of matching a complete variant name. Only fields belonging to active constraints are documented in the target task description. Return a feasible unvisited customer when continuing the route; check `feasibility_check(node)` before returning it. Return `destination_node` only to end the current route.
\end{promptbox}
\begin{lstlisting}[style=PaperCode]
from collections.abc import Callable

import numpy as np


def select_next_node(
    instance: dict,
    current_node: int,
    destination_node: int,
    unvisited_nodes: np.ndarray,
    feasibility_check: Callable[[int], bool],
) -> int | None:
    """Select one feasible next node for the active route.

    Args:
        instance: State and candidate metrics described in the task. The
            active_constraints list identifies primitive constraint membership.
        current_node: int, last node on the active route.
        destination_node: int, depot/destination. Returning it ends the route.
        unvisited_nodes: np.ndarray[int], shape (m,).
        feasibility_check: callable that checks whether a node can be appended.

    Returns:
        A feasible node from unvisited_nodes, or destination_node to end the route.
    """
    pass
\end{lstlisting}

The FJSP example describes the release-date source task. The interface takes the current instance, the partial schedule, and the admissible actions.

\begin{promptbox}{FJSP task description}
Design a constructive operation-machine dispatch heuristic for a target with active primitives `'r'`. `instance['active_constraints']` lists the active lowercase primitives. Compose behavior from membership in this list; do not hard-code complete variant names. Each action is `(operation_id, machine_id)`. `actions` already contains only ready and eligible choices; return one member and rank choices for makespan quality. Do not reimplement readiness. Always available: `operations`, `job_operations`, and the partial schedule in `state`, including `scheduled`, `machine_sequences`, and `job_previous`. Active `r` provides `release_dates`.
\end{promptbox}
\begin{lstlisting}[style=PaperCode]
import math
import numpy as np

def select_action(
    instance: dict,
    state: dict,
    actions: list[tuple[int, int]],
) -> tuple[int, int]:
    """Choose one ready operation-machine action.

    Args:
        instance: Current FJSP instance with active constraint IDs.
        state: Partial schedule and predecessor information.
        actions: Ready and eligible (operation_id, machine_id) pairs.

    Returns:
        One pair contained in actions.
    """
    pass
\end{lstlisting}

\subsection{Descriptions for active attributes}
\label{app:prompt_attributes}

Only the descriptions of active constraints are appended to the task prompt. Each family has four such descriptions.

\begin{promptbox}{VRP constraints}
O: open routes

B: interleavable linehaul and backhaul service; a route starting with a pickup starts empty and contains only pickups
Linehaul/backhaul membership is available in the instance state.

TW: Time-window candidate metrics provide normalized `arrival_ratio`, `waiting_ratio`, `slack_ratio`, and `finish_ratio`; route state provides `current_time_ratio` and `time_window_slack_ratio`.

L: Route-distance state provides `route_distance`, `route_limit`, and `remaining_route_budget_ratio`.
\end{promptbox}

\begin{promptbox}{FJSP constraints}
R: Active `r` provides `release_dates`.

A: Active `a` provides `assembly_predecessors`.

S: Active `s` provides `num_setup_families`, `setup_families`, and `setup_times`.

T: Active `t` provides `transport_times`.
\end{promptbox}

\subsection{Generation instructions}
\label{app:prompt_instructions}

Each proposal uses one of the following instructions together with its selected parents and search context.

\begin{promptbox}{Mechanism composition $\mathrm{c}_1$}
Identify one concrete decision mechanism in each parent (for example a ranking signal, look-ahead, or route-ending rule). Create a child that composes only compatible mechanisms. Do not paste two complete score functions together; explain the composed mechanism briefly in the algorithm description.
\end{promptbox}

\begin{promptbox}{Conflict resolution $\mathrm{c}_2$}
Identify a concrete conflict between the decision biases of Parent A and Parent B. Create a compact conditional or hierarchical rule that resolves the conflict from the documented runtime state. The result must be one coherent heuristic, not an average of two arbitrary scores.
\end{promptbox}

\begin{promptbox}{Exploratory crossover $\mathrm{c}_3$}
The parents occupy different regions of the measured root-performance Pareto set. Design a coherent candidate that improves the under-covered target dimension named in the search context using a genuinely different decision principle or look-ahead structure. Do not diversify by renaming variables or perturbing constants.
\end{promptbox}

The mutation instruction below uses one parent and its source evaluation.
\begin{promptbox}{Single-parent mutation $\mathrm{m}_1$}
Keep the parent algorithm as the backbone. Use the task description and source evaluation to identify a decision mechanism to improve. Revise that mechanism while preserving the parent's useful structure and the documented function interface.
\end{promptbox}
\clearpage
\endgroup

\clearpage
\section{Representative evolved heuristics}
\label{app:heuristics}

We present two detailed heuristics from the frozen sets produced by the MECo searches, selected for their program structure rather than test performance. The listings preserve their executable code and omit only the main function's interface documentation.

\begingroup
\lstset{style=PaperCode}

\subsection{VRP example}
\label{app:heuristic_routing}

The VRP example is the third selected member from the first search. It combines feasibility filtering with a score based on travel distance and temporal features. After selecting a customer, an additional route-budget check can terminate the route instead.

\begin{lstlisting}[style=PaperCode,linerange={1-4,18-61}]
from collections.abc import Callable
import numpy as np

def select_next_node(instance: dict, current_node: int, destination_node: int, unvisited_nodes: np.ndarray, feasibility_check: Callable[[int], bool]) -> int | None:
    active_constraints = instance.get('active_constraints', [])
    candidate_metrics = instance.get('candidate_metrics', {})
    distance_matrix = instance.get('dismatrix', {})
    remaining_capacity = instance.get('remaining_capacity', float('inf'))
    route_distance = instance.get('route_distance', 0)
    route_limit = instance.get('route_limit', float('inf'))
    remaining_route_budget_ratio = instance.get('remaining_route_budget_ratio', 1.0)
    feasible_nodes = []
    for node in unvisited_nodes:
        node = int(node)
        if not feasibility_check(node):
            continue
        metrics = candidate_metrics.get(node)
        if metrics is None:
            continue
        if 'capacity' in active_constraints:
            demand = metrics['demand']
            remaining_after = metrics['remaining_after']
            if demand > remaining_capacity:
                continue
            if 'l' in active_constraints and demand > 0 and (remaining_after <= 0):
                continue
        if 'tw' in active_constraints and (not metrics.get('feasible', True)):
            continue
        feasible_nodes.append(node)
    if not feasible_nodes:
        return int(destination_node)
    scores = {}
    for node in feasible_nodes:
        dist_to_node = distance_matrix[current_node][node]
        metrics = candidate_metrics[node]
        wait_ratio = metrics.get('waiting_ratio', 0.0)
        finish_ratio = metrics.get('finish_ratio', 0.0)
        arrival_ratio = metrics.get('arrival_ratio', 0.0)
        slack_ratio = metrics.get('slack_ratio', 0.0)
        score = dist_to_node + wait_ratio + finish_ratio + arrival_ratio * 0.5
        if 'tw' in active_constraints and (not metrics.get('feasible', True)):
            score += 1000
        scores[node] = score
    selected_node = min(scores, key=scores.get)
    dist_next = distance_matrix[current_node][selected_node]
    if route_distance + dist_next > route_limit * remaining_route_budget_ratio:
        return int(destination_node)
    return int(selected_node)
\end{lstlisting}

\clearpage
\subsection{FJSP example}
\label{app:heuristic_fjsp}

The FJSP example is the fourth selected member from the third search. It estimates start times from release dates, predecessor completion, transportation delays, and machine availability. Its ranking adds an extra setup penalty to estimated completion time when setups are active, favoring assignments with smaller setup costs. The environment continues to enforce action feasibility.

\begin{lstlisting}[style=PaperCode,linerange={1-4,15-59}]
import math
import numpy as np

def select_action(instance: dict, state: dict, actions: list[tuple[int, int]]) -> tuple[int, int]:
    if not actions:
        raise ValueError('No ready action')
    operations = {int(operation['id']): operation for operation in instance['operations']}
    active = set(instance.get('active_constraints', ()))

    def duration(operation_id: int, machine: int) -> int:
        """Fetch the processing duration for a given operation on a specified machine."""
        for alternative in operations[operation_id]['alternatives']:
            if int(alternative['machine']) == int(machine):
                return int(alternative['duration'])
        return 10 ** 12

    def compute_start_time(operation_id: int, machine: int) -> int:
        """Compute the start time for an operation on a machine considering release and transport times."""
        start_time = int(instance['release_dates'][int(operations[operation_id]['job'])]) if 'r' in active else 0
        previous = state['job_previous'].get(operation_id)
        if previous is not None:
            previous_record = state['scheduled'][previous]
            start_time = max(start_time, int(previous_record['end']))
            if 't' in active:
                start_time = max(start_time, int(previous_record['end']) + int(instance['transport_times'][int(previous_record['machine'])][machine]))
        sequence = state['machine_sequences'][machine]
        last_operation = sequence[-1] if sequence else None
        machine_ready_time = 0 if last_operation is None else int(state['scheduled'][last_operation]['end'])
        return max(start_time, machine_ready_time)

    def compute_setup_time(operation_id: int, machine: int) -> int:
        """Calculate setup time based on the last operation's family."""
        if 's' in active:
            families = instance['setup_families']
            last_operation = state['machine_sequences'][machine][-1] if state['machine_sequences'][machine] else None
            previous_family = int(instance['num_setup_families']) if last_operation is None else int(families[last_operation])
            return int(instance['setup_times'][machine][previous_family][int(families[operation_id])])
        return 0

    def compute_completion(action: tuple[int, int]) -> int:
        """Calculate the total completion time for an action."""
        (operation_id, machine) = map(int, action)
        start_time = compute_start_time(operation_id, machine)
        setup_time = compute_setup_time(operation_id, machine)
        start_time += setup_time
        processing_duration = duration(operation_id, machine)
        return start_time + processing_duration
    best_action = min(actions, key=lambda action: compute_completion(action) + (compute_setup_time(action[0], action[1]) if 's' in active else 0))
    return tuple(map(int, best_action))
\end{lstlisting}
\endgroup

\clearpage
\section{Licenses}
\label{app:licenses}

The licenses and URLs of baselines are listed in Table~\ref{tab:asset_licenses}.

\begin{table}[ht]
\centering
\normalsize
\caption{Licenses and URLs of the baseline implementations used in this work.}
\label{tab:asset_licenses}

\begin{tabular*}{\linewidth}{@{\extracolsep{\fill}}lll@{}}
\toprule
Method & License & URL \\
\midrule
EoH & BSD 2-Clause & \url{https://github.com/Optima-CityU/LLM4AD} \\
ReEvo & BSD 2-Clause & \url{https://github.com/ai4co/reevo} \\
MCTS-AHD & BSD 2-Clause & \url{https://github.com/Optima-CityU/LLM4AD} \\
EoH-S & MIT & \url{https://github.com/FeiLiu36/EoH-S} \\
MEoH & BSD 2-Clause & \url{https://github.com/Optima-CityU/LLM4AD} \\
HSEvo & MIT & \url{https://github.com/datphamvn/HSEvo} \\
MoH & MIT & \url{https://github.com/yiding-s/MoH} \\
EMO-STA & Apache 2.0 & \url{https://github.com/alperengozeten/emo} \\
\bottomrule
\end{tabular*}
\end{table}

\end{document}